\pdfoutput=1

\documentclass[11pt]{article}

\usepackage[final]{acl}
\usepackage{multirow}
\usepackage{graphicx}     
\usepackage{subcaption} 
\usepackage[bottom]{footmisc}
\usepackage{hyperref}
\usepackage{tablefootnote} 
\usepackage{caption} 
\usepackage{times}
\usepackage{booktabs} 
\usepackage{xcolor}
\usepackage{latexsym}
\usepackage{amssymb}
\usepackage[T1]{fontenc}
\usepackage{booktabs}
\usepackage{tabularx}

\usepackage[utf8]{inputenc}

\usepackage{microtype}
\usepackage{colortbl} 
\usepackage{inconsolata}

\usepackage{graphicx}
\usepackage{amsmath}
\usepackage{url}
\title{
Neural Incompatibility: The Unbridgeable Gap of Cross-Scale Parametric Knowledge Transfer in Large Language Models
}

\author{Yuqiao Tan$^{1,2}$, Shizhu He$^{1,2}$
\thanks{Corresponding author}, \textbf{Kang Liu}$^{1,2}$, \textbf{Jun Zhao}$^{1,2}$  \\
    $^1$ The Key Laboratory of Cognition and Decision Intelligence for Complex Systems, \\
    Institute of Automation, Chinese Academy of Sciences, Beijing, China \\
    $^2$ School of Artificial Intelligence, University of Chinese Academy of Sciences, Beijing, China \\
  {tanyuqiao2025@ia.ac.cn} {\{shizhu.he, jzhao, kliu\}@nlpr.ia.ac.cn} \\
  }

\begin{document}
\maketitle
\begin{abstract}
Large Language Models (LLMs) offer a transparent brain with accessible parameters that encode extensive knowledge, which can be analyzed, located and transferred.    
Consequently, a key research challenge is to transcend traditional knowledge transfer paradigms rooted in symbolic language and achieve genuine Parametric Knowledge Transfer (PKT).
Significantly, exploring effective methods for transferring knowledge across LLMs of different scales through parameters presents an intriguing and valuable research direction.
In this paper, we first demonstrate \textbf{Alignment} in parametric space is the fundamental  prerequisite to achieve successful cross-scale PKT. We redefine the previously explored knowledge transfer as Post-Align PKT (PostPKT), which utilizes extracted parameters for LoRA initialization and requires subsequent fine-tune for alignment.
Hence, to reduce cost for further fine-tuning, we introduce a novel Pre-Align PKT (PrePKT) paradigm and propose a solution called
\textbf{LaTen} (\textbf{L}oc\textbf{a}te-\textbf{T}h\textbf{e}n-Alig\textbf{n}) that aligns the parametric spaces of LLMs across scales only using several training steps without following training. 
Comprehensive experiments on four benchmarks demonstrate that both PostPKT and PrePKT face challenges in achieving consistently stable transfer.   
Through in-depth analysis, we identify \textbf{Neural Incompatibility} as the ethological and parametric structural differences between LLMs of varying scales, presenting fundamental challenges to achieving effective PKT. These findings provide fresh insights into the parametric architectures of LLMs and highlight promising directions for future research on efficient PKT.
Our code is available at \url{https://github.com/Trae1ounG/Neural_Incompatibility}.
\end{abstract}

\section{Introduction}

\begin{figure}[t]
\centerline{\includegraphics[width=0.5\textwidth]{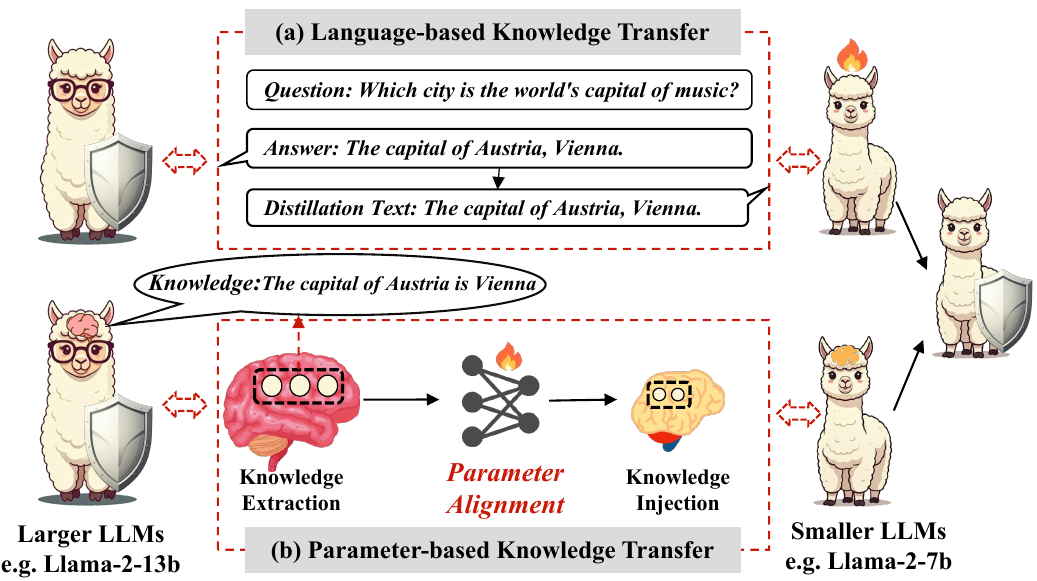}}
\caption{
Different paradigms of knowledge transfer between cross-scale LLMs. Compared to human-like symbolic knowledge transfer based on language (as shown in (a)), we aspire for LLMs to achieve more efficient knowledge transfer leveraging knowledgeable parameter (as illustrated in (b)).
}
\label{fig:intro}
\end{figure}

Human beings have non-transparent thoughts without inherited memory, requiring them to learn through communication in language-based settings. 
Based on this language-based knowledge transfer paradigm (Figure~\ref{fig:intro}(a)), Large Language Models (LLMs) have acquired a wealth of knowledge and abilities to understand and solve general tasks, with a massive amount of knowledge encoded in their parameters during pretraining on an extensive corpus~\cite{achiam2023gpt,brown2020language, ouyang2022training}. However,  unlike the unknowable and opaque nature of the human brain, the accessible parameters and information flow of LLMs (e.g. Llama~\cite{touvron2023llama}) function as a transparent brain that directly encodes factual knowledge, which can be systematically analyzed, precisely located and effectively transferred.

Existing studies~\cite{,geva2020transformer, wang2022interpretability, yu2024neuron} have made significant progress in interpreting knowledge localization and information flow in the transparent brain of LLMs, allowing possible knowledge manipulation. 
Model merging~\cite{wortsman2022model, matena2022merging, yu2024language} combines models with different capabilities to create a multitask capable model by merging weights.
Although this approach performs parametric knowledge merge, it is limited to models of the same scale and the same checkpoint. 
In real-world scenarios, it is more common for larger and smaller LLMs to form natural paired.
Compared to smaller LLMs $M_s$, larger LLMs $M_l$ encapsulate more world knowledge due to they contain larger parameter scales and extended training processes, making them intuitively appear as more extensively trained versions of smaller LLMs. Building on this insight, the key research question becomes: \textit{Can knowledge be effectively transferred from larger LLMs to smaller ones through parameters?} 

We define this challenge as \textbf{Parametric Knowledge Transfer} (\textbf{PKT}) in cross-scale LLMs.
Then, We first demonstrate through simple experiments that achieving \textbf{Alignment} in the parametric space is a prerequisite for successful cross-scale PKT. 
Notably, PKT comprises three key stages: 1) \textbf{Knowledge Extraction}, which extracts task-related knowledge from $M_l$; 2) \textbf{Parameter Alignment}, which aligns the extracted knowledge with $M_s$; and 3) \textbf{Knowledge Injection}, which performs the final parameters integration.
Existing research on PKT performs alignment after knowledge injection by further training, which we refer to this paradigm as Post-Align PKT (PostPKT).
PostPKT uses the extracted parameters to initialize certain modules (e.g.  Low-Rank Adaptation (LoRA)~\cite{hu2021lora}) for injection while holding the overall parameter unchanged. 
 \textsc{Seeking}~\cite{zhong2023seeking} addresses PostPKT by employing sensitivity-based knowledge location combined with LoRA-driven injection, followed by additional fine-tuning for alignment. Despite achieving improved performance after training for several epochs on 1,000 examples, this approach incurs high alignment costs. As a result, it is not only expensive but also unable to directly enhance the model's performance.

To reduce cost for further fine-tuning, we introduce a novel paradigm of PKT, which takes parameter alignment before injection called Pre-Align PKT (PrePKT). In this paradigm, we aspire to directly enhance LLMs ability after injection. To achieve this goal, we propose \textbf{LaTen} (\textbf{L}oc\textbf{a}te-\textbf{T}h\textbf{e}n-Alig\textbf{n}) to facilitate the alignment of parametric spaces in LLMs across different scales (Figure~\ref{fig:intro}(b)).
Specifically, LaTen uses neuron-level attribution~\cite{yu2024neuron} to address discrepancies in layer number and identify the most informative neurons for transfer in both feed-forward networks (FFNs) and multi-head self-attention (MHSA) modules. To perform dimensionality reduction, we use a simple MLP-based hypernetwork, which learns to map the parameter space of $\Theta_{l}$ to $\Theta_{s}$ by training on a small subset for alignment ($\textless$ 100). By decoding just one seed sample with the larger model, task-related parameters can be identified and projected into the target parametric space, enabling immediate improvements in downstream task performance.

Our experiments focus on PostPKT and PrePKT in three benchmark categories: world knowledge, mathematical reasoning, and code generation, using Llama-2-based models~\cite{touvron2023llama}. For PostPKT, we compare \textsc{Seeking} with PiSSA~\cite{meng2024pissaprincipalsingularvalues} which decomposes the original parameters of $\Theta_s$ to derive LoRA. Our results show that LoRA parameters derived from larger $M_l$ are less effective compared to those derived from the model $M_s$ itself.
Concurrently, although our proposed LaTen demonstrates promising performance, it still faces challenges in achieving consistently stable PrePKT.
 Through in-depth analysis, we identify \textbf{Neural Incompatibility} as the ethological and parametric structural differences between cross-scale LLMs which are similar to the cross-species neural mechanism~\cite{lu2024detecting, wang2025chimpanzee}, presenting fundamental challenges to achieve optimal parametric knowledge transfer. These findings offer novel insights into the parametric structures of LLMs and suggest directions for future research on efficient PKT.
Our main contributions are summarized as follows:

\begin{itemize}
    \item We are the first to comprehensively define and explore parametric knowledge transfer between cross-scale LLMs.
    \item We identify the importance of alignment and systematically study parametric knowledge transfer from Pre-Align and Post-Align paradigms.
    \item We propose a novel method Locate-Then-Align to first try to solve Pre-Align challenge, which leverages neuron attribution and hypernetwork techniques to execute alignment with minimal training data achieves promissing performance.  
    \item Comprehensive quantitative and qualitative assessments have highlighted the neural incompatibility as a key challenge arising from ethological and parametric structural differences in cross-scale LLMs.
\end{itemize}

\section{Related Work}
\subsection{Location of Parametric Knowledge}
Large language models (LLMs) encode vast amounts of knowledge in their parameter space through pre-training on large-scale corpora. Consequently, numerous studies have focused on identifying where knowledge is stored in language models, particularly in "neurons"~\cite{song2024does,tang2024language,niu2024does,chen2024analyzing,chen2024journey}. ~\citet{dai2021knowledge} first introduced the term “knowledge neuron” referring to the specific medium within the model that stores knowledge. Their work demonstrated that the factual knowledge encoded in a model’s parameters could be modified by manipulating these neurons. Building on this, ~\citet{meng2022locating} refined the process of identifying knowledge in LLMs using causal tracing, showing that FFN layers in the middle blocks of the model are critical for encoding factual knowledge.
To address the computational overhead of these methods, ~\citet{yu2024neuron} proposed a static attribution approach inspired by the logit lens~\cite{nostalgebraist2020logitlens}, enabling the identification of important neurons with reduced computational and memory requirements. Based on these advancements, we adopt neurons as the fundamental units for our work and apply a modified neuron attribution method adapted from~\cite{yu2024neuron} to achieve knowledge extraction.
\subsection{Manipulation of Parametric Knowledge}
With the recognition of how knowledge stored in model parameters, exist research has sought to execute diverse operations on these parameters, aiming to manipulate the implicit knowledge. Knowledge editing aims to update the parameters related to specific knowledge in a model without affecting its other capabilities~\cite{de2021editing,mitchell2021fast, meng2022locating, meng2022mass}. 
We categorize editing as fine-grained knowledge manipulation, whereas its counterpart, coarse-grained manipulation, encompasses tasks such as model merging~\cite{jin2022dataless, yu2024language,bowen2024beyond}. Model merging typically starts with models of the same scale, or even identical checkpoints, to combine multiple models with different capabilities into a single, multitask model. A series of studies regard the delta parameter as task vector~\cite{ilharco2022editing, zhang2023composing, huang2024chat}, which can be leveraged through arithmetic operations such as addition and subtraction to acquire or forget specific skills, demonstrating strong generalization capabilities. DyPRAG~\cite{tan2025dynamic} transforms symbolic documents into parametric knowledge using a hypernetwork. However, these approaches are either limited to individual models or require maintaining the same scale and even identical checkpoints. Moreover, they do not explore how parametric knowledge can be transferred across models of different scales.
\subsection{Transfer of Parametric Knowledge}
Large language models  can achieve explicit knowledge transfer through language or logits, known as knowledge distillation~\cite{hinton2015distilling}. However, this approach overlooks the rich parametric knowledge encoded within the model's weights. Existing methods for parametric knowledge transfer primarily focus on model merging, directly combining parameters from models of same scale. However, little attention has been given to transferring knowledge across models of different scales. While inference-time proxy-tuning techniques~\cite{liu2024tuning, wu2024cross} allow smaller models to influence larger ones by adjusting output logits, they do not directly leverage the implicit parametric knowledge. Recent work \textsc{Seeking}~\cite{zhong2023seeking} has explored knowledge transfer between models of different scales. It leverages sensitivity-based extraction and LoRA-driven injection, followed by Post-Alignment using training. In this study, we conduct a comprehensive analysis of both PostPKT and our proposed PrePKT to investigate strategies for achieving efficient and effective PKT.
\section{Challenge of Pre-Align Parametric Knowledge Transfer}
\textsc{Seeking}~\cite{zhong2023seeking}, designed to address Post-align PKT, leverages delta parameters $\Delta\Theta$ extracted from $M_l$ for LoRA initialization, demonstrating advantages over random initialization. Since the overall parameters remain unchanged after injection, we refer to the subsequent fine-tuning process as \textbf{post-alignment}. This process, which aligns the LoRA parameters from $M_l$ with the original ones, plays a critical role in enhancing PostPKT.
However, the post-alignment process is expensive and requires training on large-scale data. A more intuitive and cost-effective approach is to directly inject a certain form of $\Delta \Theta$ into the original parameters, thereby immediately enhancing task-specific knowledge without the need for additional training.
In this section, we try to solve this without alignment to determine its significance in this new context.
We adopt several straightforward and commonly used methods. However, we find that all of these unaligned transfer methods fail. Through analysis, we identify key statistical factors contributing to this failure, underscoring the critical importance of \textbf{pre-alignment} in our newly proposed Pre-Align PKT paradigm.
\label{sec:challenge}
\subsection{Analysis Setup}
We first illustrate how Transformer~\cite{vaswani2017attention} works and denote several symbols, detailed in Appendix~\ref{app:transformer}.  
Due to the parametric space mismatch between $M_l$ and $M_s$, the achievement of knowledge transfer hinges on discrepancy dimension and layer number problem. For example, chat version of Llama-2-7b has a layer number $L$ of 32, hidden dimension $d$ of 4096 and FFN neuron numbers $N$ of 11008, while 40, 5120 and 13824 for chat version of Llama-2-13b, respectively. To reach the conclusion, we employ several basic approaches, relying solely on pre-selected layers and standard dimensionality reduction techniques. For layer selection, we propose three methods: \textsc{Top-}$L_s$, \textsc{Bottom-}$L_s$ and \textsc{Random-}$L_s$. Given a smaller LLM $M_s$ with $L_s$ layers, these methods select the top, bottom, or random $L_s$ layers from $M_l$, respectively. To match hidden dimension $d_l$ and $d_s$, we use standard dimensionality reduction techniques \textsc{PCA}~\cite{abdi2010principal}, \textsc{Whitening}~\cite{bert} and a learning-based \textsc{Embedding Transform}. For the numbers of FFN neuron $N_l$ and $N_s$, \textsc{Random} and \textsc{Importance} are selected. Meanwhile, we consider sensitivity-based knowledge localization method \textsc{Seeking}~\cite{zhong2023seeking} as the strongest among the evaluated baselines. Notably, we directly using the extracted $\Delta\Theta_{\textrm{extract}}$ as delta parameters, i.e. $\Theta_s^\prime=\Theta_s+\Delta\Theta_{\textrm{extract}}$. Details on these methods can be found in Appendix~\ref{app:clarify}.
\subsection{Results of Unaligned Baselines}
\begin{figure}[t]
\centerline{\includegraphics[width=0.5\textwidth]{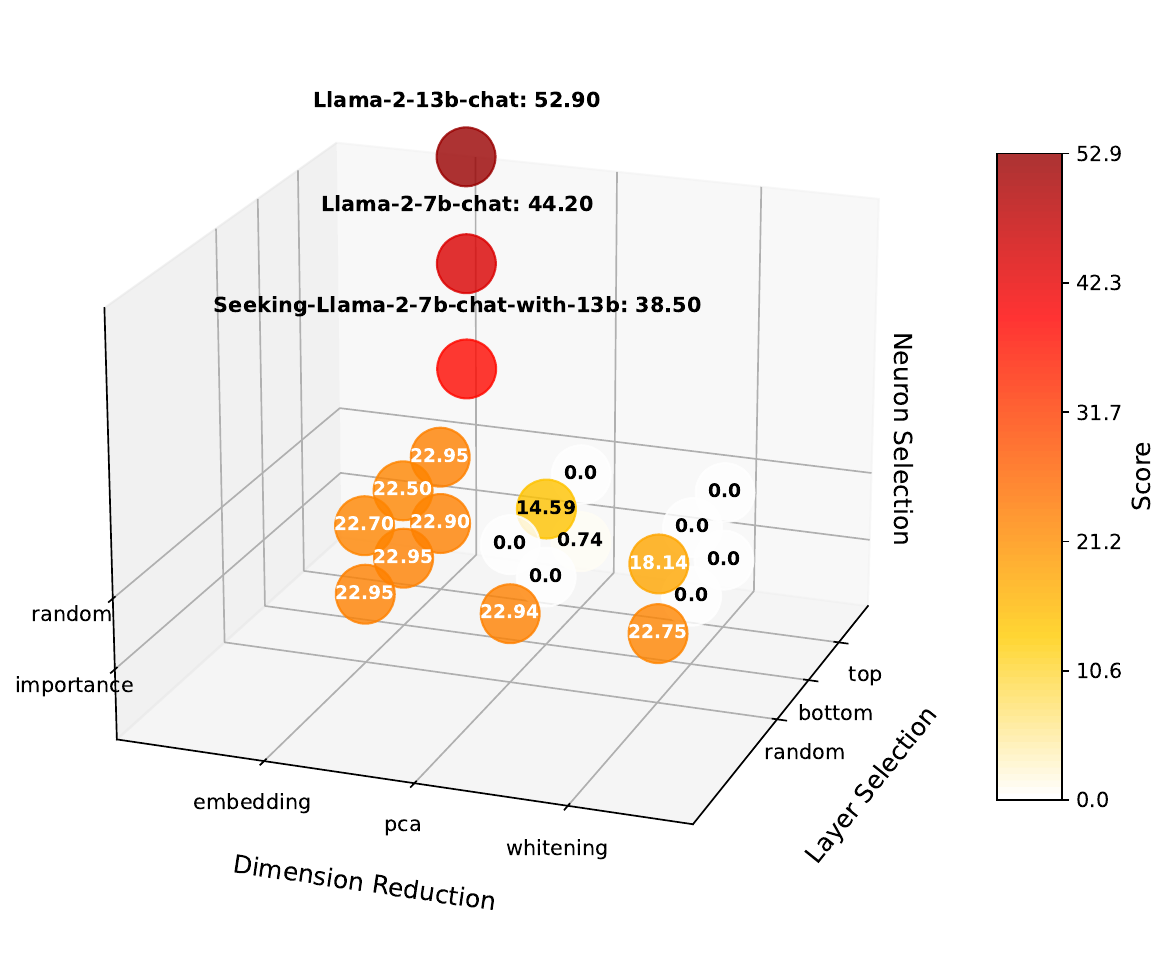}}
\caption{
Performance of different baseline methods in MMLU.
}
\label{fig:baseline}
\end{figure}

The results are presented in Figure~\ref{fig:baseline} using MMLU benchmark~\cite{hendrycks2021measuring}. The Llama-2-13b-Chat and Llama-2-7b-Chat models score 52.90 and 44.20, respectively. All unaligned transfer methods significantly impair the model's ability to perform specific tasks, with some cases resulting in nearly zero functionality. For instance, when using the \textsc{Importance} method for neuron selection and the \textsc{Embedding Transform} for dimensionality reduction, several approaches achieve a score of 22.95 with a degradation of 21.25. 
Notably, \textsc{Seeking} also causes a significant performance drop of 5.70 compared to the original model.

We begin our statistical analysis of the delta parameter ranges and observe that the delta parameters from regular supervised fine-tuning (SFT) exhibit a distinct pattern compared to those from the \textsc{Seeking} method. Specifically, the delta parameter ranges for normal SFT are smaller (within 0.002, as shown in Figures~\ref{fig:Llama2_7b_parameter_change_range} and~\ref{fig:Llama2_13b_parameter_change_range} aligns with~\cite{yu2024language}), while those from \textsc{Seeking} are larger (exceeding 0.005 in Figure~\ref{fig:Llama2_seeking_gsm_parameter_change_range}), with some even surpassing 1. Consequently, directly injecting these large range parameters can significantly alter the parameter distribution (e.g. potentially flipping $\theta_i$ from positive to negative).

Furthermore, we find that unaligned delta parameters exhibit low parametric similarity to the original weights, indicating an absence of useful task-related information in LoRA initialization. In contrast, well-aligned SFT delta parameters demonstrate significantly higher similarity to the original weights. We conduct further in-depth analysis in Section~\ref{sec:analysis}. Based on these observations, we conclude that alignment is a crucial factor for achieving PKT. Therefore, implementing pre-alignment in the parametric space during PrePKT presents a significant challenge.

\section{LaTen: Locate-Then-Align for Pre-Align Parametric Knowledge Transfer}
Based on the observations and conclusions in Section~\ref{sec:challenge}, we propose a novel Locate-Then-Align (LaTen) method to solve PrePKT and achieve parametric space alignment, while preventing to disrupt the injected model and without additional training. Specifically, we utilize static neuron-level attribution~\cite{yu2024neuron} method to locate knowledgeable parameters for transferring, according to both FFN and MHSA neurons contain task-related parametric knowledge~\cite{geva2023dissecting, chen2024yes}. The neuron-level attribution method is detailed in Appendix~\ref{app:neuron}. 
\subsection{Overall Parametric Knowledge Transfer Definition}
Considering a smaller LLM $M_s$ and a larger LLM $M_l$, intially parameterized by $\Theta_{s}$ and $\Theta_{l}$, respectively.  For a specific task $\mathcal{T}$, corresponding to a training dataset $D^{\mathcal{T}}_{\textrm{train}}=\{(x^{\mathcal{T}}_i,y^{\mathcal{T}}_i)\}^Q_{i=1}$ comprising $Q$ input-output instances and a extract dataset $D^{\mathcal{T}}_{\textrm{extract}}$ and a alignment dataset $D^{\mathcal{T}}_{\textrm{align}}$. The overall goal is to extract delta parameters $\Delta\Theta^{\mathcal{T}}_{\textrm{extract}}$ from $\Theta_l$ based on $D^{\mathcal{T}}_{\textrm{extract}}$, then align to $\Theta_s$ to obtain $\Delta\Theta^\mathcal{T}_{\textrm{algin}}$:
\begin{equation}
\Delta\Theta^\mathcal{T}_{\textrm{extract}}=\textnormal{Extract}(\Theta_l;\Theta_s;D^{\mathcal{T}}_{\textrm{extract}})
\end{equation}
\begin{equation}
     \Delta\Theta^{\mathcal{T}}_{\textrm{align}} = \textnormal{Align}(\Delta\Theta^\mathcal{T}_{\textrm{extarct}}; D^{\mathcal{T}}_{\textrm{align}} \;\textrm{or} \; D^\mathcal{T}_{\textrm{train}})
\end{equation}
where $\textnormal{Extract}(\cdot)$ representing the logic for parameter extraction and $\textnormal{Align}(\cdot)$ encapsulating the alignment process for cross-scale PKT. Notably, $D^{\mathcal{T}}_{\textrm{align}}$ only uses in PrePKT, while PostPKE trains for alignment on larger $D^{\mathcal{T}}_{\textrm{train}}$. Knowledge injection will execute to merge the delta parameters into $\Theta_s$:

\begin{equation}
    \Theta^\mathcal{T}_s=
    \begin{cases}
        \Theta_s - \Delta\Theta^\mathcal{T}_{\textrm{extract}} + BA, & \hfill \textnormal{for PostPKT} \\
        \Theta_s + \Delta\Theta^\mathcal{T}_{\textrm{align}}, & \hfill \textnormal{for PrePKT} \\
    \end{cases}
\end{equation}
where in PostPKT, the LoRA $BA$ matrices are initialized using $\Delta\Theta^\mathcal{T}_{\textrm{extract}}$ by SVD~\cite{golub1971singular}, and the injection is performed before the $\textnormal{Align}(\cdot)$ while keeping the overall weight unchanged. In contrast, our proposed PrePKT first performs $\textnormal{Align}(\cdot)$ and then executes injection. By incorporating the well-aligned parameters, the model ability on task $\mathcal{T}$ is directly enhanced.
\subsection{Neuron-level Localization for Knowledge Extraction}
Existing studies have demonstrated that neurons in FFN and MHSA of Transformer serve as fundamental units for storing knowledge or specific skills~\cite{geva2020transformer, dai2021knowledge}. These neurons can be leveraged to modify the model's behavior, such as editing~\cite{meng2022locating, meng2022mass} in single model, which have not been fully explored in a couple of models of different scales.

We use the neuron-level attribution method~\cite{yu2024neuron} to locate task-related useful neurons for knowledge extraction. As detailed in Appendix~\ref{app:neuron}, we use this method to gain importance score of each neuron vector $v$, denoted as $Imp(v)$, can be computed by measuring the change in the output distribution of a predicted token $t$. Since an answer $y$ typically consists of $T$ tokens, we only choose the last useful token $t_T$ for attribution score. This process produces a score matrix $S$ with dimensions $S^{\textrm{FFN}}\in\mathbb{R}^{L \times N}$ for the FFN and $S^{\textrm{MHSA}}\in\mathbb{R}^{L \times d}$ for the MHSA, where each element $S_{i,j}$ represents the importance score of neuron $v$ in the $i$th layer at the $j$th position. To perform layer selection, we sum $S_{i,j}$ over layers to get $S^{\textrm{FFN}}_{\textrm{layer}}, S^{\textrm{MHSA}}_{\textrm{layer}} \in \mathbb{R}^{L}$ then choosing the top-$L_s$ layers for $M_l$. To align with neuron number of $M_s$, we select the top-$c$ neurons $I_l$ in layer $l$ by sorting $S_{l,\cdot}$ in descending order, where $c$ equals $N_s$ or $d_s$ for FFN or MHSA, repectively. Based on $I_l$, we then extract the corresponding key and value neurons to obtain the unaligned delta parameters $\Delta\Theta_{\textrm{extract}}^\mathcal{T} \in \mathbb{R}^{N_s \times d_l}$ for FFN or $\mathbb{R}^{d_s \times d_l}$ for MHSA.

\subsection{Parameter Alignment with Knowledge Injection}
$\textnormal{Align}(\cdot)$  as a prerequisite before knowledge injection.
Given the absence of explicit parameter-level correspondence between $M_s$ and $M_l$, we employ traditional language modeling loss for alignment optimization.
Our training framework operates as follows: Based on each training instance $(x^{\mathcal{T}}_i,y^\mathcal{T}_i) \in D^\mathcal{T}_{\textrm{extract}}$, we first extract the raw delta parameters $\Delta\Theta^\mathcal{T}_{\textrm{extract}}$. These parameters are then processed through a lightweight hypernetwork (implemented as a two-layer MLP with ReLU activation) to obtain dimension reduced parameters $\Delta\Theta^\mathcal{T}_{\textrm{align}}$. The target model parameters are subsequently injected via $\Theta^\mathcal{T}_s=\Theta_s+\Delta\Theta^{\mathcal{T}}_{\textrm{align}}$. 
During alignment, we randomly sample $P$ instances from the alignment dataset $D^\mathcal{T}_{\textrm{algin}}$ to compute the language modeling loss, updating the hypernetwork for effective parameter alignment. During evaluation, we randomly sample one unseen instance, process extracted parameters through trained hypernetwork for final alignment, and apply the aligned parameters to the target model.
\begin{table*}[htbp]
\centering
\begin{tabular}{lccccc}
\toprule
Models & MMLU & GSM8K  & HumanEval & MBPP \\ 
\midrule
Llama-2-7B-Chat       & 44.20  & 16.07 & 14.05  & 17.80  \\ 
Llama-2-13B-Chat      & 52.90  & 20.55 & 18.75 & 19.20   \\ 
\rowcolor{gray!20} \multicolumn{5}{l}{\textit{\# Post-Align Parametric Knowledge Transfer}}\\
Post-Align on $D^\mathcal{T}_{\textrm{train}}$ (=1000): \\
\quad -Random Initialization. & 49.73 & 26.51 & 14.22 & 15.60\\
\quad -Seeking + 13B Param. & 49.60 & 28.23 & 15.44 & 20.60\\ 
\quad -PiSSA Initialization.   & \textbf{49.77}& \textbf{29.32} & \textbf{16.26} & \textbf{21.40}\\
Post-Align on $D^{\mathcal{T}}_{\textrm{align}}$ ($\textless$ 100): \\
\quad -Random Initialization. & 44.20 & 16.35& 14.02 & 18.20\\
\quad -Seeking + 13B Param. & 44.20 & 14.78& 14.63 & 18.60\\
\rowcolor{gray!20} \multicolumn{5}{l}{\textit{\# Pre-Align Parametric Knowledge Transfer}}\\
\quad -Seeking + Unaligned 13B Param wo Train. & 38.50 & 7.28 & 0.61 & 0.00\\
Pre-Align on $D^{\mathcal{T}}_{\textrm{align}}$ ($\textless$ 100): \\
\quad -\textbf{LaTen + Pre-Aligned 13B Param.} & \textbf{44.40} & \textbf{20.47} & \textbf{14.63} & 18.20\\
\bottomrule
\end{tabular}
\caption{Results for Post-Align and Pre-Align parametric knowledge transfer.}
\label{tab:overall}
\end{table*}
\section{Experiments}

\subsection{Experimental Setup}
\noindent \textbf{Datasets and Pre-trained Backbones.}
We select MMLU~\cite{hendrycks2021measuring} to evaluate the professional knowledge of models.
 We choose GSM8K~\cite{cobbe2021training} for evaluating mathematical reasoning ability. For code generation, HumanEval~\cite{chen2021evaluating} and MBPP~\cite{austin2021program} are adopted for estimation. For pre-trained LLMs, we use Llama 2~\cite{touvron2023llama} to conduct task-related PKT.

\noindent \textbf{Evaluation Metrics.}
We calculate  zero-shot accuracy for GSM8K and MMLU, pass@1 for HumanEval and MBPP.

\noindent \textbf{Implementation Details.} 
Since chosen benchmark require instruction following ability which not include in base version, we use chat version in remain experiments. 
To execute PKT, we randomly sample three non-overlapping subsets from the original training dataset (we utilize python examples from~\cite{luo2023wizardcoder} as training set for HumanEval\footnote{\url{https://huggingface.co/datasets/nickrosh/Evol-Instruct-Code-80k-v1}}): an extract set of size 32, an align set of size 80, and a training set of size 1000 (except MBPP). However, not all examples in the alignment set are used for alignment in LaTen. Details of the count are provided in Table~\ref{tab:datasets_prepkt}. We denote the $D^{\mathcal{T}}_{\textrm{align}}$ as actual examples used for parameter alignment in the following section. Additional implementation details and experimental settings can be found in \ref{app:implement}.
\subsection{Experimental Results}

\noindent \textbf{Main Results for Post-Align PKT.} 
\textsc{Seeking} is the first to investigate the transferability of cross-scale LLMs and demonstrated that using delta parameters as LoRA initialization can significantly enhance the ability of $M_s$ to acquire task-specific knowledge. Since this paradigm keeps the overall parameters unchanged at the beginning, it can be considered as a variant of LoRA initialized from larger LLMs. Therefore, we are curious whether the optimal LoRA initialization originates from another model or from the model itself?

To explore this, we compare it with a self-derived LoRA approach: PiSSA~\cite{meng2024pissaprincipalsingularvalues}, which applies singular value decomposition (SVD) to separate LoRA parameters and residual components from the original model parameters. As shown in the upper section of Table~\ref{tab:overall}, \textsc{Seeking} outperforms random Gaussian initialized LoRA in most scenarios. For instance, the \textsc{seeking} method achieves an improvement of 5.0 compared to Gaussian initialization in MBPP. However, \textsc{Seeking} shows even lower performance in MMLU task, demonstrating its instability. Notably,  PiSSA consistently achieves higher performance across all benchmarks compared to \textsc{seeking}, with the delta parameters derived from the same model providing an additional 0.72 boost in performance on average.

These results suggest that the \textsc{seeking} method, which relies on parameters from a different LLM, requires more effort but leads to suboptimal performance. We hypothesize that this is due to the incompatibility of the delta parameters from $M_l$ compared to $M_s$ itself.
\begin{table}[t]
\centering
\small
\begin{tabularx}{\columnwidth}{Xcc} 
\toprule
Models & HumanEval & MBPP \\ 
\midrule
Llama-2-7B       & 14.05 & 17.80  \\ 
\quad -PiSSA Initialization & \textbf{16.26} & \textbf{21.40} \\
\rowcolor{gray!20} \multicolumn{3}{l}{\textit{\# Post-Align PKT from Llama-2-13B}}\\
Llama-2-13B      & 18.75 & 19.20 \\ 
\quad -Seeking + 13B Param. & 15.44 & 20.60 \\ 
\rowcolor{gray!20} \multicolumn{3}{l}{\textit{\# Post-Align PKT from  WizardCoder-13B-Python}}\\
WizardCoder-13B-Python & 56.71 & 41.60\\
\quad -Seeking + 13B Param. & 15.04  & 19.80\\
\rowcolor{gray!20} \multicolumn{3}{l}{\textit{\# Post-Align PKT from CodeLlama-13B-Python}}\\
CodeLlama-13B-Python & 47.56 & 37.80 \\
\quad -Seeking + 13B Param. & 16.05 & \textbf{21.40}\\
\bottomrule
\end{tabularx}
\caption{Results for Post-Align PKT from different larger LLMs in code generation.}
\label{tab:stronger_code}
\end{table}

\noindent \textbf{Main Results for Pre-Align PKT.} 
In Section~\ref{sec:challenge}, we explored various methods to extract unaligned $\Delta\Theta^{\mathcal{T}}_{\textrm{extract}}$  for direct application to the parameters. However, none of these methods yielded satisfactory results. Instead, they degraded the model original capabilities. We also show more comprehensive results in Table~\ref{tab:overall} to further demonstrate our finding of the importance of alignment.

As shown in the lower part of Table~\ref{tab:overall}, our proposed LaTen achieves strong performance across all benchmarks. For instance, LaTen improves by 4.40 on GSM8K and achieves an average improvement of 1.86 across four datasets compared to the base model $M_s$. 
We also employ a more equitable setup to compare the PostPKT and PrePKT paradigms, where PostPKT is restricted to performing post-align operations solely on the align dataset $D^\mathcal{T}_{\textrm{align}}$ (\textless 100).
In this setting, LaTen outperforms models with random initialization. Additionally, when compared to the PostPKT \textsc{Seeking} approach with the same computational cost for alignment, LaTen demonstrates superior performance in most settings, particularly on GSM8K, where it achieves an improvement of 5.69. These results highlight the potential of PrePKT paradigm and LaTen in effectively solving it.

It is worth emphasizing that our proposed LaTen achieves powerful performance with only a few steps of parameter alignment. However, this phenomenon is a double-edged sword. Identifying the best checkpoint requires multiple experiments, and the parameter space alignment process does not exhibit a straightforward minimum point, making it more challenging to optimize compared to traditional language-based transfer methods. Therefore, exploring stable PrePKT methods is valuable for future research.

\begin{table}[t]
\centering
\small
\begin{tabularx}{\columnwidth}{Xcc} 
\toprule
Models & HumanEval & MBPP \\ 
\midrule
Llama-2-7B       & 14.05 & 17.80  \\ 
\rowcolor{gray!20} \multicolumn{3}{l}{\textit{\# Pre-Align PKT from Llama-2-13B}}\\
Llama-2-13B      & 18.75 & 19.20 \\ 
\quad --LaTen + 13B Param. & \textbf{14.63} & 18.20 \\ 
\rowcolor{gray!20} \multicolumn{3}{l}{\textit{\# Pre-Align PKT from WizardCoder-13B-Python}}\\
WizardCoder-13B-Python & 56.71 & 41.60\\
\quad -LaTen + 13B Param. & 14.02  & \textbf{18.60}\\
\rowcolor{gray!20} \multicolumn{3}{l}{\textit{\# Pre-Align PKT from CodeLlama-13B-Python}}\\
CodeLlama-13B-Python & 47.56 & 37.80 \\
\quad -LaTen + 13B Param. & 14.02 & 17.80\\
\bottomrule
\end{tabularx}
\caption{Results for Pre-Align PKT from different larger LLMs in code generation.}
\label{tab:stronger_code_pre}
\end{table}

\subsection{Analysis}
\noindent \textbf{Can Stronger $M_l$ Transfer Richer Knowledge?}
In the above experiment, we observed that using the parameters derived from larger LLMs $M_l$ via \textsc{Seeking} to initialize LoRA generally outperforms random initialization in most cases. However, it still falls short compared to using LoRA initialized with parameters derived from the model itself~\cite{meng2024pissaprincipalsingularvalues}. To further investigate whether the parameters extracted from $M_l$ carry useful information, an intuitive hypothesis is that parameters extracted from $M_l$ which specialized in a specific task should be more useful for PKT.

To test this hypothesis, we employ WizardCoder-13B-Python~\cite{luo2023wizardcoder} and CodeLlama-13B-Python~\cite{rozière2024codellamaopenfoundation} as $M_l$ for comparison. Both WizardCoder and CodeLlama are fine-tuned on code-specific datasets to enhance their code generation capabilities, outperforming Llama-2-13B in this domain. However, as shown in Table~\ref{tab:stronger_code}, under identical fine-tuning conditions in PostPKT, initializing LoRA parameters with WizardCoder-13B unexpectedly resulted in worse task performance. For example, on the HumanEval (MBPP) task, performance decreased by 0.4 (0.8) compared to Llama-2-13B. Although CodeLlama-13B demonstrated improvements over Llama-2-13B, it still falls short of matching the performance of PiSSA. We also examine mathematical generation task in Table~\ref{tab:stronger_math} which shows the similar results in PostPKT.
For PrePKT, the results also do not support our hypothesis, shown in Table~\ref{tab:stronger_code_pre}, 

Based on the above experiments, we propose that LLMs of different parameter scales inherently exhibit incompatibility in parameters, which makes ideal PKT largely coincidental. We call this \textbf{Neuron Incompatibility} and further explore the reasons behind in following discussion.

\label{sec:analysis}
\begin{figure}[t]
    \centering
    \begin{subfigure}[b]{0.23\textwidth}
        \centering
        \includegraphics[width=\textwidth]{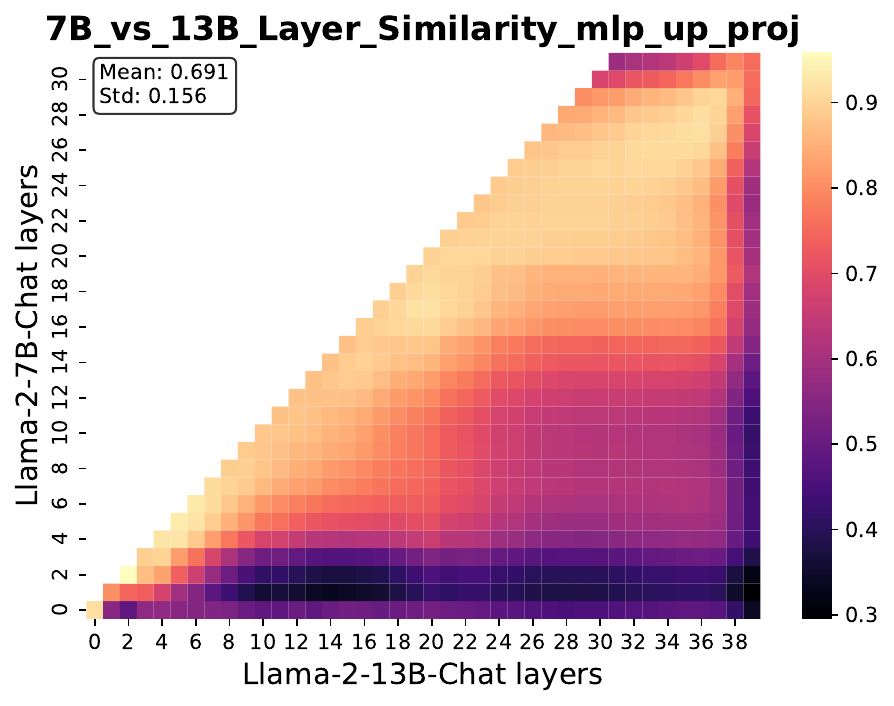}
    \end{subfigure}
    \hfill
    \begin{subfigure}[b]{0.23\textwidth}
        \centering
        \includegraphics[width=\textwidth]{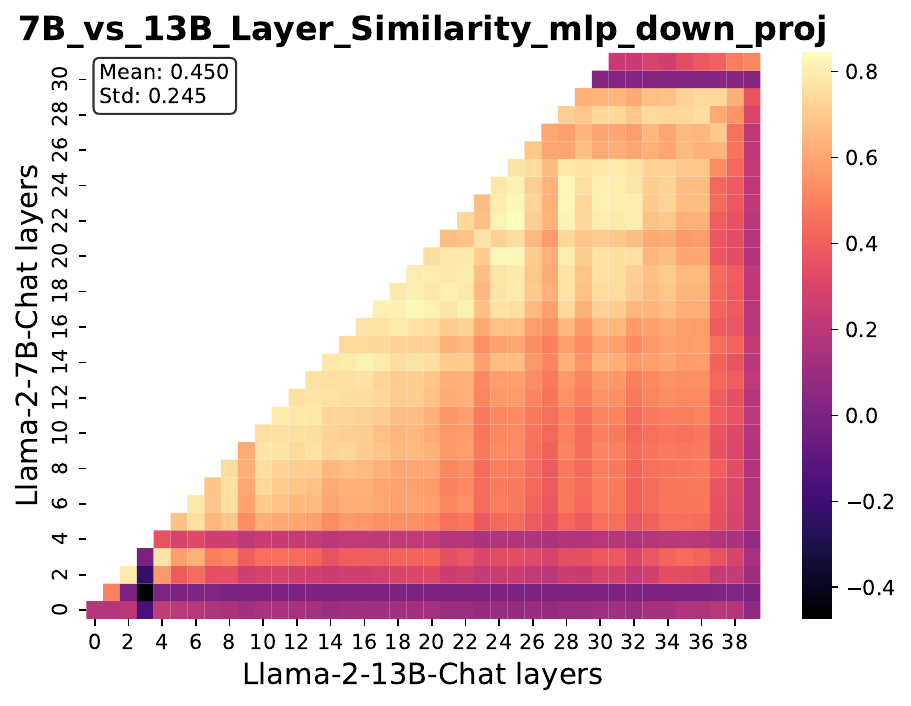}
    \end{subfigure}
    \\
    \hfill
    \begin{subfigure}[b]{0.23\textwidth}
        \centering
        \includegraphics[width=\textwidth]{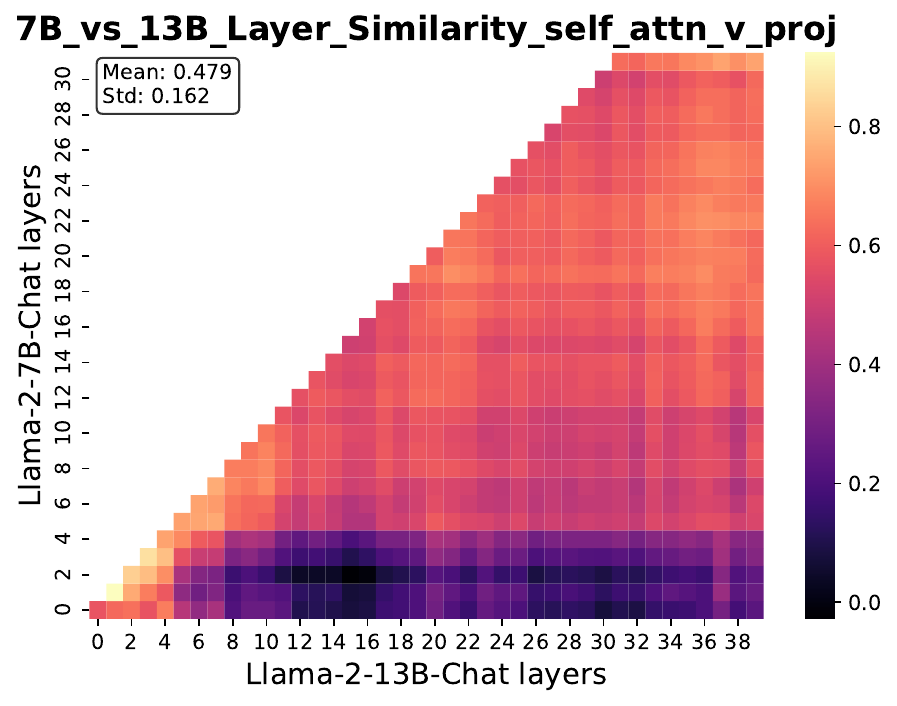}
    \end{subfigure}
    \hfill
    \begin{subfigure}[b]{0.23\textwidth}
        \centering
        \includegraphics[width=\textwidth]{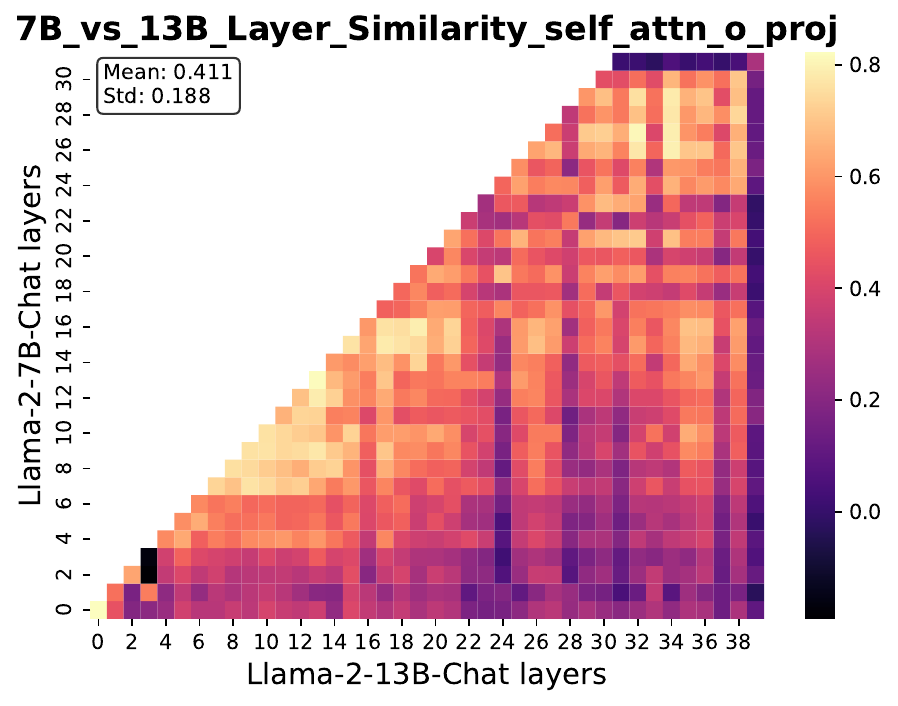}
    \end{subfigure}
    \caption{Representation Similarity Comparison Results between LLMs.}
    \label{fig:representation_similarity}
\end{figure}
\noindent \textbf{Ethological  Similarity between Cross-Scale LLMs.}
To further analyze why both PostPKT and PrePKT perform suboptimally, we utilize Centered Kernel Alignment (CKA)~\cite{kornblith2019similarityneuralnetworkrepresentations}, a method based on the Hilbert-Schmidt Independence Criterion (HSIC), to compute the similarity between feature representations in neural networks. This metric assesses the similarity in behaviors between the two models, which can be interpreted as the ethological similarity of LLMs. We compute the ethological similarity between Llama-2-7B and Llama-2-13B across the up-proj, down-proj, v-proj, and o-proj modules.

As shown in Figure~\ref{fig:representation_similarity}, the similarity between 7B and 13B is notably low, especially in the MHSA module which plays the most important part for integrating information~\cite{elhage2021mathematical}. Interestingly, the up-proj layers demonstrate higher similarity, likely because they function as key memories, capturing specific input patterns~\cite{geva2020transformer}, which tend to be consistent across models. The weak similarity between $M_l$ and $M_s$ also explains why LoRA derived from the same model performs better, as it aligns more closely with the model's intrinsic behavior. 
We identify that the weak ethological similarity between cross-scale LLMs is one of the key factors contributing to neural incompatibility, making ideal parametric knowledge transfer success difficult to achieve.

\begin{figure}[htbp]
    \centering
    \begin{subfigure}[b]{0.23\textwidth}
        \centering
        \includegraphics[width=\textwidth]{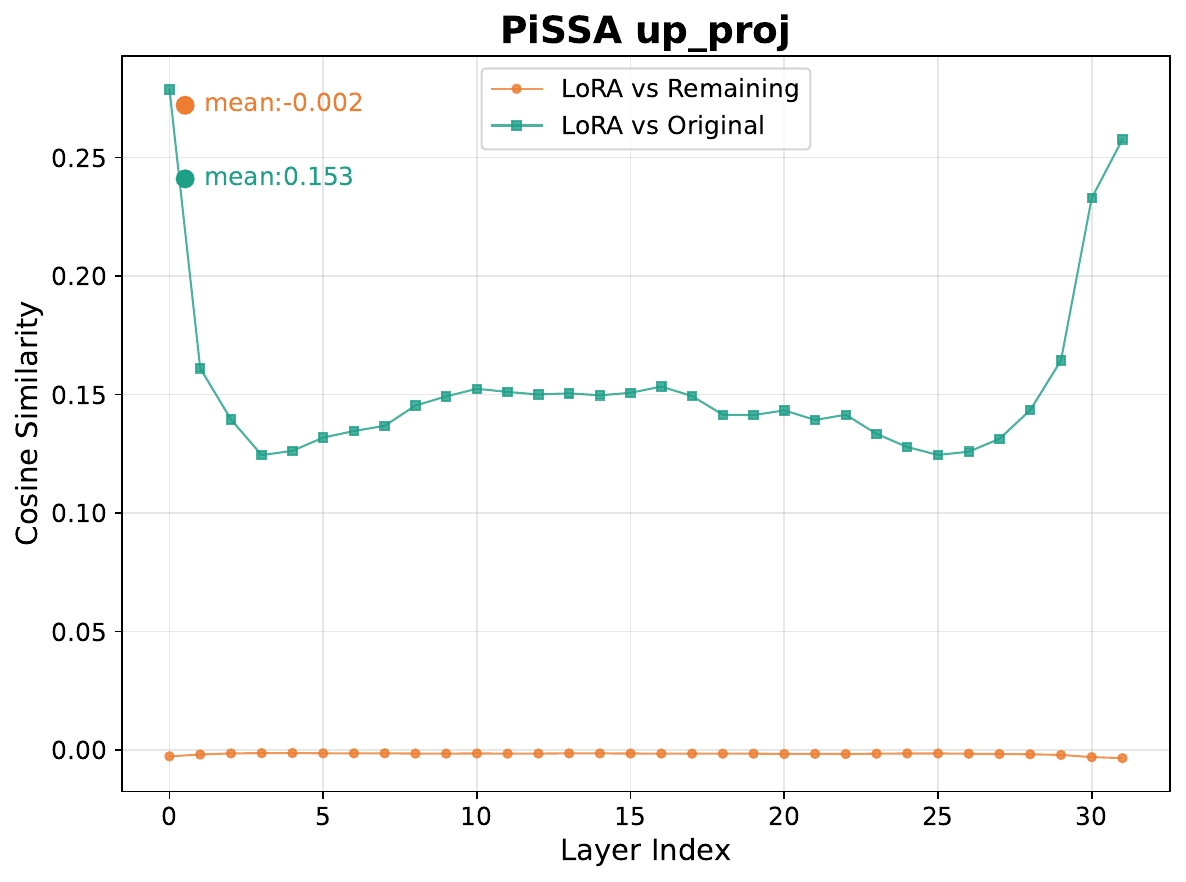}
    \end{subfigure}
    \hfill
    \begin{subfigure}[b]{0.23\textwidth}
        \centering
        \includegraphics[width=\textwidth]{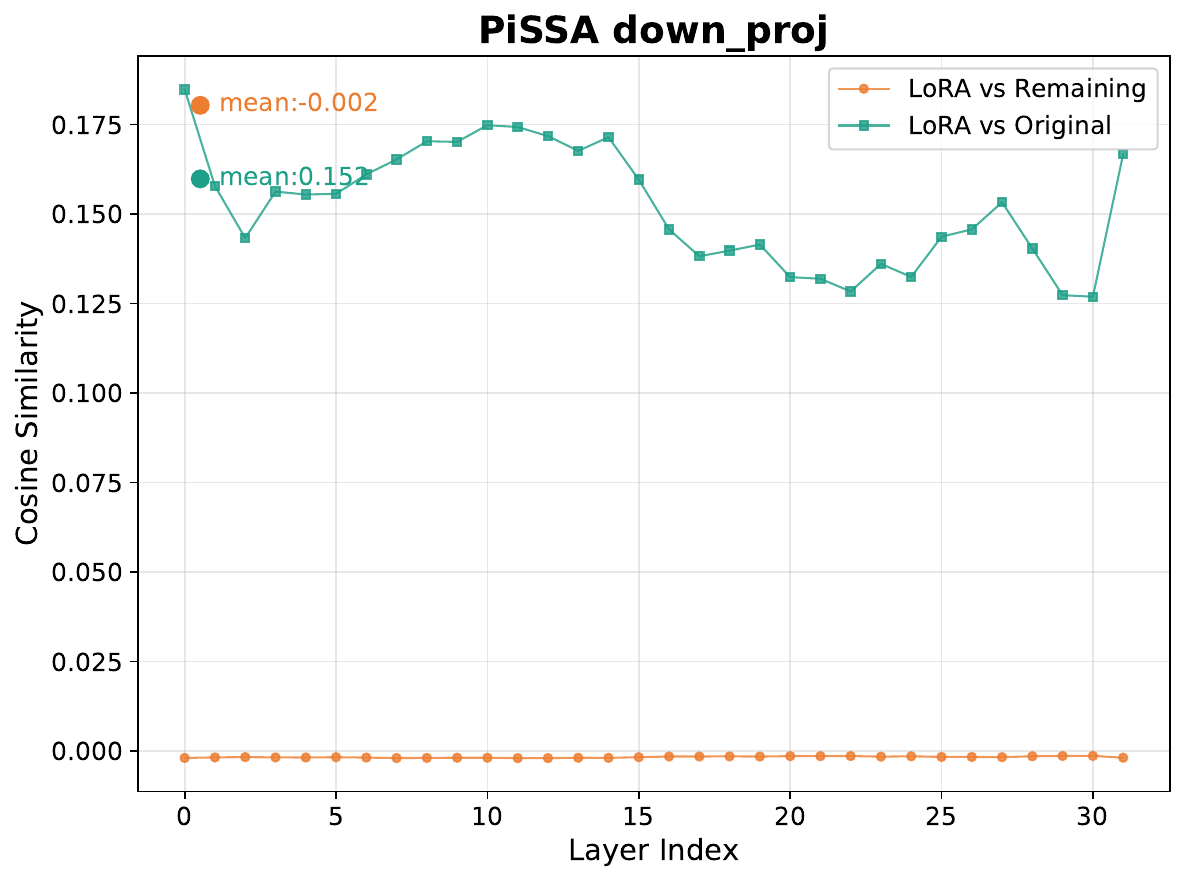}
    \end{subfigure}
    \\
    \hfill
    \begin{subfigure}[b]{0.23\textwidth}
        \centering
        \includegraphics[width=\textwidth]{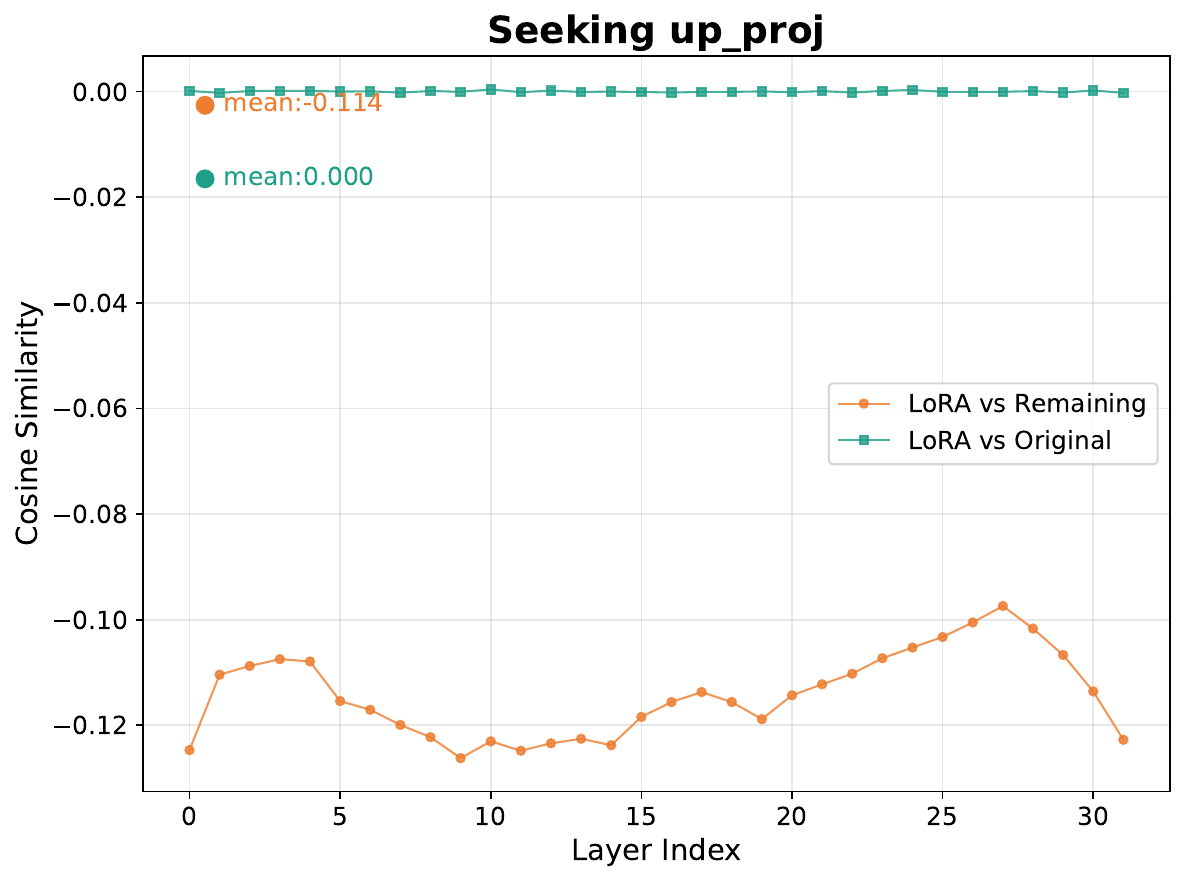}
    \end{subfigure}
    \hfill
    \begin{subfigure}[b]{0.23\textwidth}
        \centering
        \includegraphics[width=\textwidth]{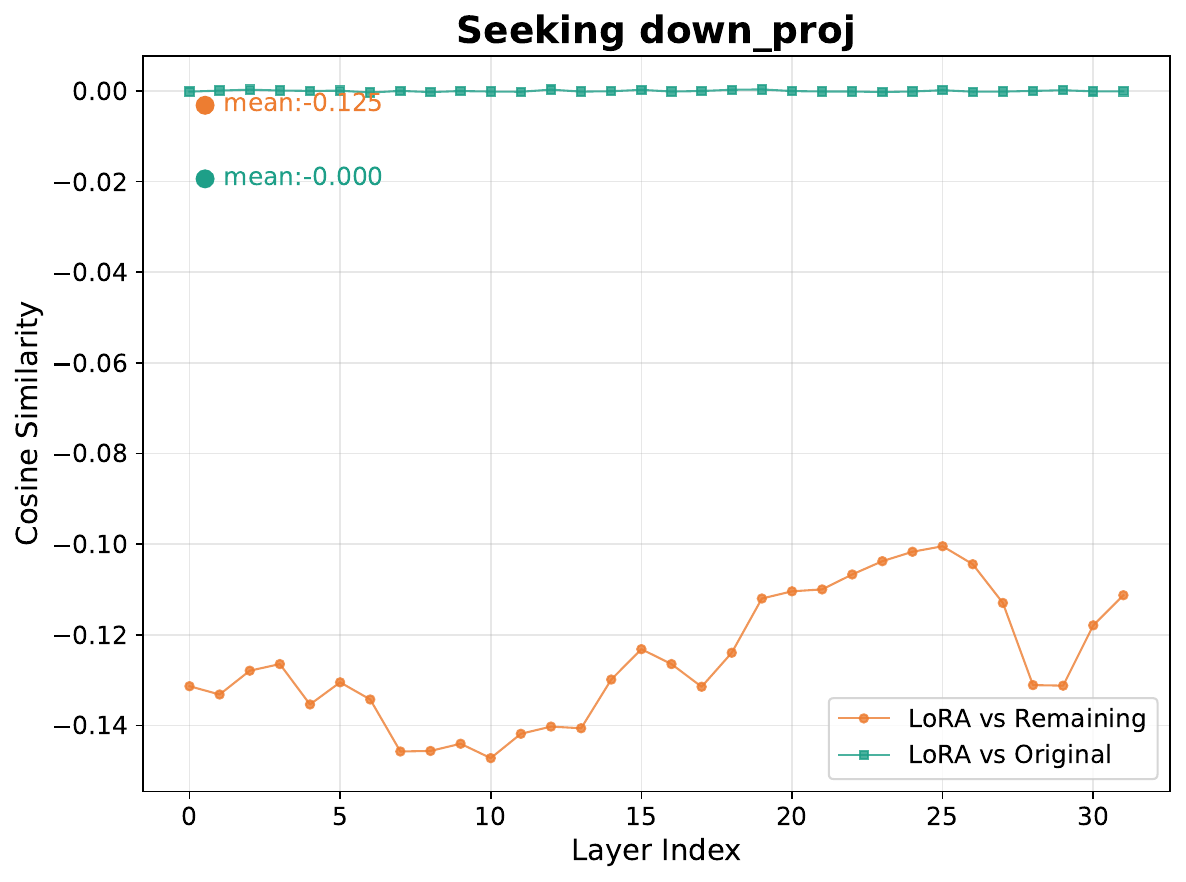}
    \end{subfigure}

    \caption{Parametric Similarity Comparison Results between LLMs in MLP Modules.}
    \label{fig:parametric_similarity}
\end{figure}

\noindent \textbf{Parametric Structural Similarity in Different Methods.}
We further conduct in-depth analysis based on parametric structural similarity to figure out whether it performs important influence on performance. As shown in Figure~\ref{fig:parametric_similarity}, we compare $W^l_{\text{LoRA}}$ (i.e. LoRA parameters in layer $l$) with both $W^l$ and $W^l_{\text{remain}}$ (i.e. $W^l-W^l_{\text{LoRA}}$) in up-proj and down-proj modules. First, the pattern of results is completely opposite between \textsc{Seeking} and PiSSA. In \textsc{Seeking}, the mean similarity between $W^l$ and $W^l_{\textrm{LoRA}}$ drops to 0, suggesting that $W^l_{\textrm{LoRA}}$ does not retain any meaningful information from $W^l$. 
This deficiency results in suboptimal performance. For comparison, PiSSA which utilizes SVD to capture important parameters for LoRA, preserves greater similarity to the original weights and establishes an orthogonal relationship with $W^l_{\text{remain}}$, making it more effective for learning new skills. Our findings indicate that parametric structural similarity plays a crucial role in further fine-tuning. Specifically, the similarity between $W_{\textrm{LoRA}}^l$ and $W^l$ significantly influences the model's ability to adapt to new tasks and execute parameter alignment. The low degree of similarity emerges as a key factor contributing to neural incompatibility. We observe the same pattern in the MHSA module (Figure~\ref{fig:parametric_similarity_mhsa}).

\section{Conclusion}
In this work, we comprehensively define and explore the feasibility of Parametric Knowledge Transfer (PKT) between cross-scale LLMs. We propose Locate-Then-Align solution to address newly raised Pre-Align PKT challenge with reduced alignment costs. Through extensive experiments on four benchmarks, we demonstrate the neural incompatibility between cross-scale LLMs reflects in low similarity in both ethological and parametric space, which pose fundamental challenges to achieve ideal PKT. Our findings offer novel insights into the parametric structures of LLMs and aim to elucidate directions for future research on efficient PKT.
\section{Limitations}
This study identifies alignment as the key factor for achieving PKT and introduces two distinct paradigms based on this insight. Although current PKT methods are somewhat effective, they still rely on language for supervision. Developing simpler and more efficient approaches that do not depend on language guidance is a promising direction for future research. Moreover, the underlying principles behind PKT's effectiveness remain unclear, and it is also uncertain why a stronger $M_l$ yields no improvement. These questions highlight important areas for further investigation. 
Furthermore, due to equipment limitations, our experiments were restricted to models between 13B and 7B. Nevertheless, the results still provide meaningful evidence to support our conclusions. In the future, expanding experiments to larger-scale LLMs would be a worthwhile and necessary direction for exploration.

\bibliography{custom}

\newpage
\appendix
\section{Background}
\subsection{Transformer}
\label{app:transformer}
Transformer-based language models~\cite{vaswani2017attention} are at the center of state-of-the-art natural language processing~\cite{devlin2018bert, brown2020language} and have become the most popular and effective architecture, even in computer vision~\cite{dosovitskiy2020image, tian2024visual}. A decoder-only Transformer is stacked with $L$ identical blocks, mainly containing a multi-head self-attention (MHSA) module and a feed-forward network (FFN) module.

Follow~\cite{yu2024neuron}, We first detail the forward pass from the input token to the final prediction. Given an input sequence $X = [t_1, t_2, \dots, t_T]$ with $T$ tokens, the model generated the next token's probability distribution $y$ over $B$ tokens in vocabulary $V$. Each $t_i$ at position $i$ starts as a word embedding $h_i^0 \in \mathbb{R}^d$ transformed by the embedding matrix $E \in \mathbb{R}^{B \times d}$. Followed by $L$ transformer layers, each layer output $h^l_i$ (layer $l$, position $i$) is the sum of the previous layer's output $h^{l-1}_i$, the FFN output $F^l_i$ and the attention output $A^l_i$:
\begin{equation}
    h^l_i = h^{l-1}_i + F^l_i + A^l_i.
\end{equation}
The final probability distribution $y$ of the next token is computed by multiplying the unembedded matrix $E_u \in \mathbb{R}^{B \times d}$ and the last position of $L$th layer output:
\begin{equation}
    y= \textnormal{softmax}(E_u h^L_T).
    \label{eq:softmax}
\end{equation}
Diving into how do the two main components work, the FFN layer's output is computed by two linear transformations with a nonlinear function $\sigma$, while the MHSA layer's output is a weighted sum over $H$ heads on $T$ positions:
\begin{equation}
    A^l_i = \sum^H_{j=1}\textnormal{ATTN}^l_j(h^{l-1}_{1}, h^{l-1}_2, \dots, h^{l-1}_T),
\end{equation}
\begin{equation}
    F^l_i = W^l_{\textrm{down}}\sigma(W^l_{\textrm{up}}(h^{l-1}_i+A^l_i)),
\end{equation}
where $W^l_{\textrm{up}} \in \mathbb{R}^{N \times d}$ and $W^l_{\textrm{down}} \in \mathbb{R}^{d \times N}$ are two linear matrices in FFN and for clarity we do not include $W^l_{\textrm{gate}}$ used in LLama-2. ~\cite{geva2020transformer} shows that FFN emulates neural memories~\cite{sukhbaatar2015end} where the $W^l_{\textrm{up}}$ corresponds to keys and $W^l_{\textrm{down}}$ to values. Then the FFN output can be transformed into a weighted sum of FFN neurons:
\begin{equation}
    \label{eq:ffn}
    F^l_i=\sum^N_{k=1}c^l_{i,k}down^l_k,
\end{equation}
\begin{equation}
    c^l_{i,k}=\sigma(up^l_k \cdot (h^{l-1}_i + A^l_i)),
\end{equation}
where $down^l_k$ denotes the $k$th column of $W^l_{\textrm{down}}$, referred  to as the FFN subvalue. The coefficient score $c^l_{i,k}$ is computed by comparing  the residual output $h^{l-1}_i+A^l_i$ with $up^l_k$, the $k$th row of $W^l_{\textrm{up}}$, referred to as the FFN subkey. Meanwhile, the MHSA output $A^l_i$ can also be expressed as the sum of individual head outputs, where each head produces a weighted sum of the value vectors across all positions:
\begin{equation}
    \label{eq:mhsa}
    A_i^l=\sum_{j=1}^H\sum_{n=1}^T\alpha_{i,j,n}^lW_{j,l}^o(W_{j,l}^vh_n^{l-1})
\end{equation}
\begin{equation}
    \alpha_{i,j,n}^l=\textnormal{softmax}(W_{j,l}^qh_i^{l-1}\cdot W_{j,l}^kh_n^{l-1})
\end{equation}
where $W_{j,l}^q$, $W_{j,l}^k$, $W_{j,l}^v$, and $W_{j,l}^o \in \mathbb{R}^{d \times {d/H}}$ represent the query, key, value, and output matrices of the $j$th attention head in the $l$th layer. The query and key matrices are used to compute the attention weight $\alpha_{i,j,n}^l$ for the $n$th position, followed by applying the softmax function across all positions. The value and output matrices then transform the input vector at the $n$th position into the corresponding value-output vector. Finally, the output of each attention head is the weighted sum of value-output vectors across all positions.

\subsection{Neuron-level Attribution}
\label{app:neuron}
Following the definition of neurons from ~\cite{yu2024neuron}, the $k$th FFN neuron is the $k$th subvalue of $W^l_{\textrm{down}}$ which is activated by its corresponding subkey $up^l_k$. To align with the definition in FFN neurons, we regard the $k$th column of $W^o_{j,l}$ as the $k$th attention subvalue (neuron) in this head, whose subkey is the $k$ th row of $W^v_{j,l}$. This design uses the position value-output $W^o_{j,l}(W^v_{j,l}h^{l-1}_n)$ as the base unit, performing an addition of $T \times H$ vectors to generate each attention output as shown in Eq.\ref{eq:mhsa}. Each vector is derived from the attention subvalue and subkey, similar to Eq.\ref{eq:ffn}.

Instead of choosing integrated gradients~\cite{sundararajan2017axiomatic, dai2021knowledge} or causal tracing~\cite{vig2020investigating, meng2022locating}, we use a static neuron-level attribution method from~\cite{yu2024neuron}
. As introduced above, the final vector $h^L_T$ used for predicting the next token is computed as a direct sum of various neuron-level vectors. Specifically, $h^l_T$ in layer $l$ can be decomposed into two components: a single neuron vector $v$ and the remaining vector $x = h^l_T - v$. The change in the output distribution, measured using the "logits lens"~\cite{nostalgebraist2020logitlens}, can then be interpreted as the importance score for the neuron $v$:
\begin{equation}
Imp(v^l) =
\begin{cases}
\begin{aligned}
&\log(p(w|v^l+h^{l-1})) \\
&-\log(p(w|h^{l-1}))
\end{aligned}
, v^l \in l\textnormal{th MHSA} \\[8pt]
\begin{aligned}
&\log(p(w|v^l+h^{l-1}+A^l)) \\
&-\log(p(w|h^{l-1}+A^l))
\end{aligned}
,  v^l \in l\textnormal{th FFN}
\end{cases}
\end{equation}
where $w$ is the prediction token and $p(w|*)$ is computed by multiplying the vector with $E_u$ (in Eq.~\ref{eq:softmax}). 

\begin{figure}[htbp]
    \centering
    \scalebox{1.0}{
    \includegraphics[width=1.00\columnwidth]{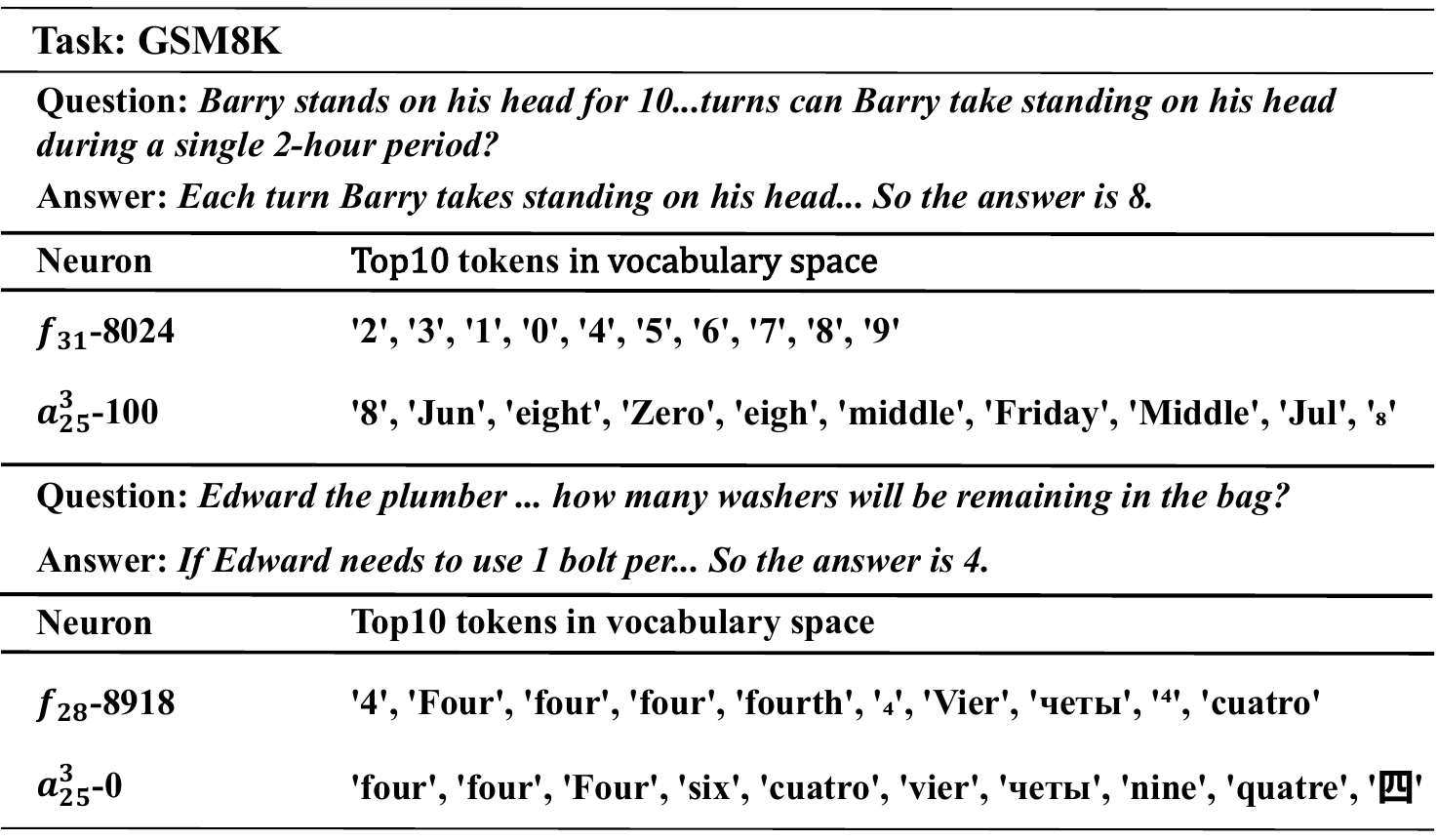}
    }
    \caption{Interpretable neuron location in GSM8K task.}
    \label{fig:neuron_attr_case_study}
\end{figure}

\noindent \textbf{Interpretable Neuron Location.} Based on our modified method~\cite{yu2024neuron}, the most influential neurons identified can be analyzed through the logits lens~\cite{nostalgebraist2020logitlens}. As illustrated in Figure~\ref{fig:neuron_attr_case_study}, we project these selected neurons into the vocabulary space, revealing that the top-10 tokens are strongly associated with both the question and answer. This approach enables effective neuron selection, preserving the most relevant neuron information for the specific task and facilitating the subsequent transfer process.

\section{Details of Parametric Knowledge Transfer Baselines}
\label{app:clarify}
\subsection{Unaligned Parametric Knowledge Transfer Baselines}
In this section, we first detail the precise implementation of \textsc{Whitening}, \textsc{Embedding Transform} and \textsc{Importance}, leaving \textsc{Seeking} in Section~\ref{app:seeking}.
\begin{itemize}
    \item \textsc{Whitening}: BERT-whitening, introduced by~\cite{bert}, provides a simple yet effective alternative to BERT-flow~\cite{li2020sentence}. It has become a widely adopted technique for dimensionality reduction while preserving key features~\cite{liao2024instance}.
    \item \textsc{Embedding Transform}: We propose an intuitive method for dimensionality reduction by learning a linear transformation $W^\prime \in \mathbb{R}^{d_l \times d_s}$ to map $E_l$ to $E_s$ using $E_s = E_l W^\prime$. Here, $E_s \in \mathbb{R}^{B \times d_s}$ and $E_l \in \mathbb{R}^{B \times d_l}$ represent the embedding matrices in $M_s$ and $M_l$, respectively. This problem is formulated as the following minimization task: 
    \begin{equation}
        \underset{W'}{\min} \|E_s - E_l W^\prime\|_F^2
    \end{equation} 
    where $\|\cdot\|_F$ denotes the Frobenius norm. After derivation, we obtain the closed-form solution: $W^\prime=(E^T_lE_l)^{-1}E^T_lE_s$ which is then applied to perform dimensionality reduction effectively.
    \item \textsc{Importance}: Following~\cite{bowen2024beyond}, we define the importance $I_i$ of the $i$th parameter $\theta_i$ using its amplitude, where $I_i = \|\theta_i\|_2^2$. Neuron sampling is performed based on average importance scores across neurons, while gradient-based scores are utilized in \textsc{Seeking}.
\end{itemize}

\subsection{Illustrate \textsc{Seeking} Method}
\label{app:seeking}
~\cite{zhong2023seeking} attempts to empirically investigate post-align parametric knowledge transfer from larger to smaller models through parametric perspective. When conducting knowledge extraction, for a given task $\mathcal{T}$, the parameter-level importance score is calculated by sensitivity~\cite{mozer1988skeletonization}:
\begin{equation}
    S^\mathcal{T}_{i,j}=\left|\theta_{i}^\top\nabla_{\theta_i}\mathcal{L}(x^\mathcal{T}_j,y^\mathcal{T}_j|\Theta)\right|
\end{equation}
where $S^\mathcal{T}_{i,j}$ represents the importance of the $i$th parameter $\theta_i$ relative to sample $j$ and the absolute value is taken for the purpose of measuring the amplitude. Then, the final score $S^\mathcal{T}_i$ for task $\mathcal{T}$ integrates the cumulative sensitivity over the sampled instances, calculated as $\sum^k_{j=1}S^\mathcal{T}_{i,j}$. Layer selection involves calculating a score for each layer by aggregating the sensitivity scores of all parameters within that layer. The layers are then ranked in descending order based on the scores, and the top $L_s$ layers are selected while preserving their original sequential order. Dimension reduction is achieved by directly extracting the sub-matrix with the highest cumulative sensitivity score:
\begin{equation}
W^{l}_{\textrm{extract}}=\arg\max_{W^{\prime}\subseteq W^{l}}\sum_{\theta_i\in\boldsymbol{W}^{\prime}}S_i.
\end{equation}
where $W^l \in \mathcal{R}^{n_l \times m_l}$ represents a matrix in $l$th layer and $W^\prime \in \mathcal{R}^{n_s \times m_s}$ is a sub-matrix in $W^l$ to match smaller model's matrix dimensions ($n_s \le n_l, m_s \le m_l$). Then aggregating $W^l_{\textrm{extract}}$ across all layers can get the final extracted parameters $\Delta\Theta_{\textrm{extract}}$.

For knowledge injection, \textsc{Seeking} utilizes LoRA~\cite{hu2021lora} as a bridge by decompose $W^l_{\textrm{extract}}$ into $U\Sigma V^T$ using Singular Value Decomposition (SVD)~\cite{golub1971singular} then transfer to $U[:,:r]\Sigma[:r,:r]V^T[:r,:]$ to match the rank $r$. In the end, \textsc{Seeking} actually provides a unaligned LoRA initialization as:
\begin{equation}
    W^{l\star} = W^l-W^l_{\textrm{extract}} + BA 
\end{equation}
where $B$ is initialized as $U[:,:r]\Sigma[:r,:r]$, and $A$ with $V^T[:r,:]$. After injection, the overall parameters remain unchanged (the SVD approximation of the LoRA loss is negligible). Subsequent post-alignment with a large amount of training data is the key process.

\subsection{Statistical Analysis}
\label{app:statistic}
\begin{figure*}[htbp]
    \centering
    \includegraphics[width=1.00\textwidth]{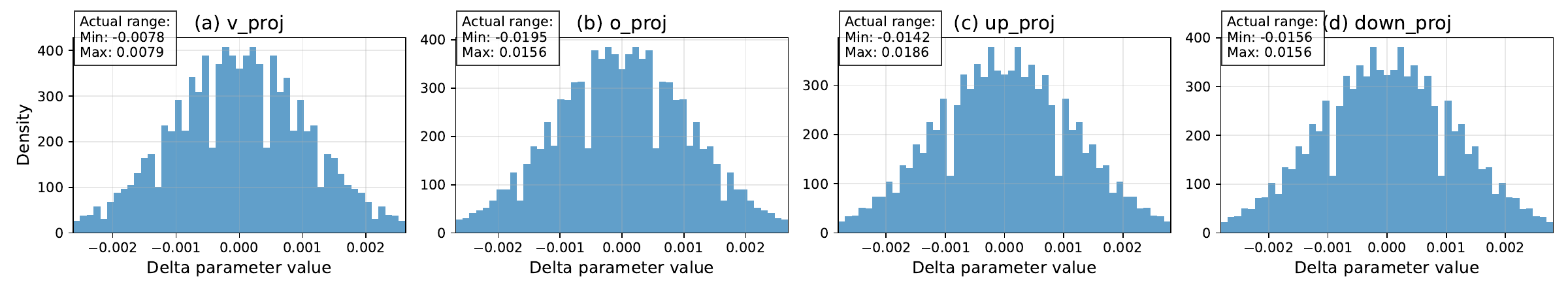}
    \caption{Delta parameter ranges between Llama-2-7b and Llama-2-7b-Chat}
    \label{fig:Llama2_7b_parameter_change_range}
\end{figure*}

\begin{figure*}[htbp]
    \centering
    \includegraphics[width=1.00\textwidth]{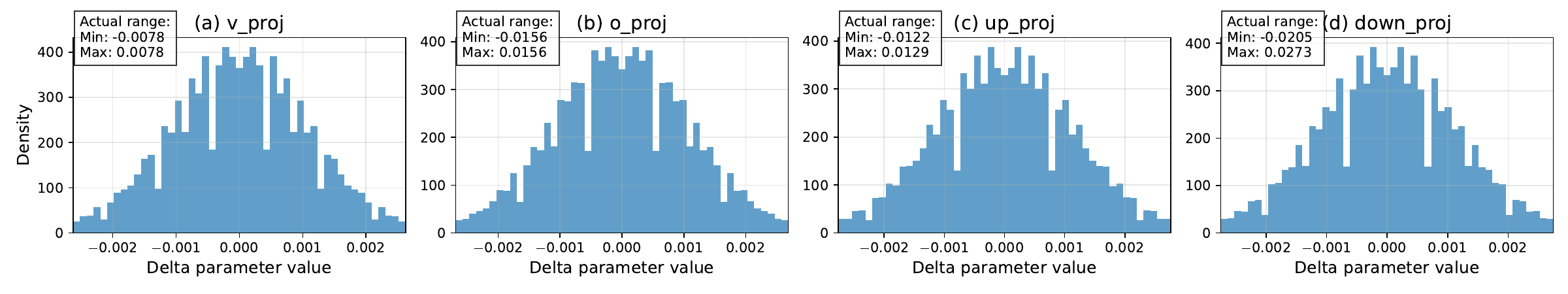}
    \caption{Delta parameter ranges between Llama-2-13b and Llama-2-13b-Chat}
    \label{fig:Llama2_13b_parameter_change_range}
\end{figure*}

\begin{figure*}[htbp]
    \centering
    \includegraphics[width=1.00\textwidth]{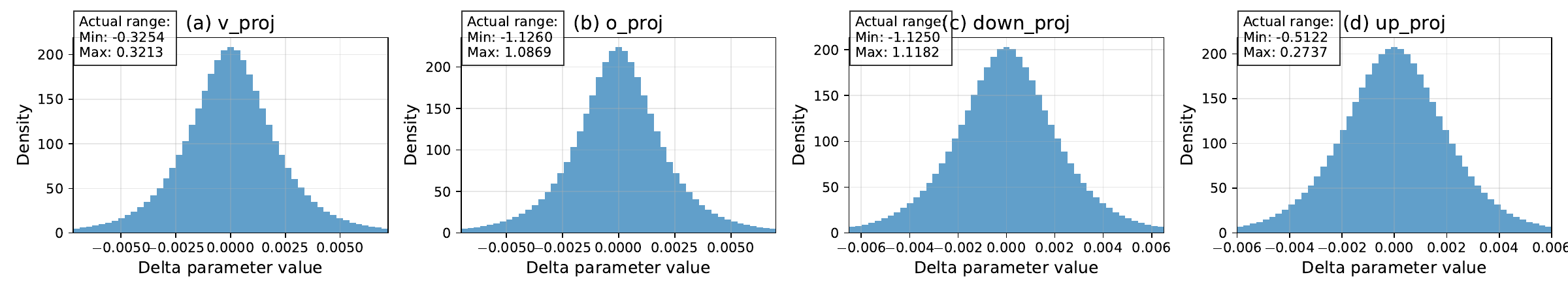}
    \caption{Delta parameter ranges from Llama-2-13b-Chat in GSM8K using \textsc{Seeking}  }
    \label{fig:Llama2_seeking_gsm_parameter_change_range}
\end{figure*}

\textbf{Range of Delta Parameters.} 
~\cite{yu2024language} finds that the SFT delta parameter ranges at a very small number (with in 0.002) and contains many redundant information that can be removed. We first derive the delta parameters from the base and chat versions of Llama-2~\cite{touvron2023llama}. The results are presented in Figures~\ref{fig:Llama2_7b_parameter_change_range} and \ref{fig:Llama2_13b_parameter_change_range} which consistent with~\cite{yu2024language}. However, using \textsc{Seeking}~\cite{zhong2023seeking} to directly extract the delta parameters from $M_l$ (e.g., Llama-2-13b) results in $\Delta\Theta_{\textrm{extract}}$ values that are poorly aligned with $M_s$, leading to a wide range of parameter differences. As shown in Figure~\ref{fig:Llama2_seeking_gsm_parameter_change_range}, these values often exceed 0.005, with a maximum of 1.12 and a minimum of -1.13. When applied directly as delta parameters, such discrepancies significantly degrade the model's performance.

\section{Detailed Experimens}

\subsection{Implementations}
\label{app:implement}
\noindent \textbf{Baseline Implementations.}
During fine-tuning, the smaller model is trained for 5 epochs with a batch size of 64 and a learning rate of 3e-4 except for HumanEval of 3e-5 and trained 3 epochs in SFT setting. Regarding LoRA, we set the rank as 16, and insert LoRA into up-proj and down-proj of FFN, v-proj and o-proj of MHSA layer. During the alignment stage, the hypernetwork is trained with a learning rate of 1e-5 and a weight decay of 0.05. We also employ the Mean Square Loss between $\Delta\Theta^\mathcal{T}_{\textrm{align}}$ and a zero tensor of the same size as a constraint. The sample size $P$ is set to 16 and we only transfer 10\% neurons in each layer. Notably, results of PostPKT are mean values from three runs with different seeds. More details about datasets are shown in Table~\ref{tab:datasets} and~\ref{tab:datasets_prepkt}. Our inference process for language models is handled  based on vLLM~\cite{kwon2023efficient}.

\noindent \textbf{Stronger $M_l$ Implementations.} 
For stronger $M_l$, we choose model which further fine-tuned on Llama-2-13B to keep comparison. We choose WizardMath-13B-V1.0\footnote{\url{https://huggingface.co/WizardLM/WizardMath-13B-V1.0}} for matheatical reasoning. We use WizardCoder-Python-13B\footnote{\url{https://huggingface.co/WizardLM/WizardCoder-Python-13B-V1.0}} and CodeLlama-13b-Python\footnote{\url{https://huggingface.co/codellama/CodeLlama-13b-Python-hf}} for code-generating task. 

\begin{table}[!htbp]
\centering
\small 
\begin{tabular}{l|c|c|c|c}
\hline
Dataset & MMLU & GSM8K & HumanEval & MBPP \\
\hline
Train Size & 1000 & 1000 & 1000 & 300 \\
\hline
lr & 3e-4 & 3e-4 & 3e-5 & 3e-4 \\
\hline
Epochs & 5 & 5 & 3 & 5 \\
\hline
\end{tabular}
\caption{Details of training datasets in PostPKT.}
\label{tab:datasets}
\end{table}

\begin{table}[!htbp]
\centering
\small 
\begin{tabular}{l|c|c|c|c}
\hline
Dataset & MMLU & GSM8K & HumanEval & MBPP \\
\hline
Align Size & 32 & 64 & 48 & 128 \\
\hline
lr & 3e-5 & 3e-5 & 3e-5 & 3e-5 \\
\hline
Steps & 2 & 4 & 3 & 8 \\
\hline
\end{tabular}
\caption{Details of alignment datasets in PrePKT.}
\label{tab:datasets_prepkt}
\end{table}

\noindent\textbf{Representation Similarity Implementations.}
In order to calculate the representation similarity, we use PyTorch implementation of Centered Kernel Alignment (CKA)\footnote{\url{https://github.com/RistoAle97/centered-kernel-alignment}}. In practice, we randomly sample 200 examples from the test set of wikitext to calculate model representations.
\subsection{More Results}
\noindent \textbf{Strong $M_l$ in Mathematical Reasoning.}
We also use WizardMath-13B-V1.0~\cite{luo2023wizardmath} as $M_l$ for comparison. WizardMath leverages Reinforcement Learning from Evol-Instruct Feedback (RLEIF) to enhance its mathematical reasoning abilities and outperforms Llama-2-13B in this domain. However, as shown in Table~\ref{tab:stronger_math}, under the same fine-tuning settings, using WizardMath-13B as the parameter source for LoRA initialization unexpectedly led to worse task performance. For instance, on the GSM task, performance dropped by 3.71 compared to Llama-2-13B and even fell 1.99 below the randomly initialized LoRA baseline. This result supports our conclusion.
\begin{table}[!htbp]
\centering
\begin{tabular}{lc}
\toprule
Models &  GSM8K \\ 
\midrule
Llama-2-7B       & 16.07    \\ 
\rowcolor{gray!20} \multicolumn{2}{l}{\textit{\# Post-Align PKT from Llama-2-13B}}\\
Llama-2-13B      & 20.55     \\ 
\quad -Seeking + 13B Param. & \textbf{28.23} \\ 
\rowcolor{gray!20} \multicolumn{2}{l}{\textit{\# Post-Align PKT from WizardMath-13B}}\\
WizardMath-13B & 53.15 \\
\quad -Seeking + 13B Param. & 24.52\\
\bottomrule
\end{tabular}
\caption{Results for implicit parametric knowledge transfer from different larger LLMs in mathematical reasoning.}
\label{tab:stronger_math}
\end{table}

\begin{figure}[t]
    \centering
    \begin{subfigure}[b]{0.23\textwidth}
        \centering
        \includegraphics[width=\textwidth]{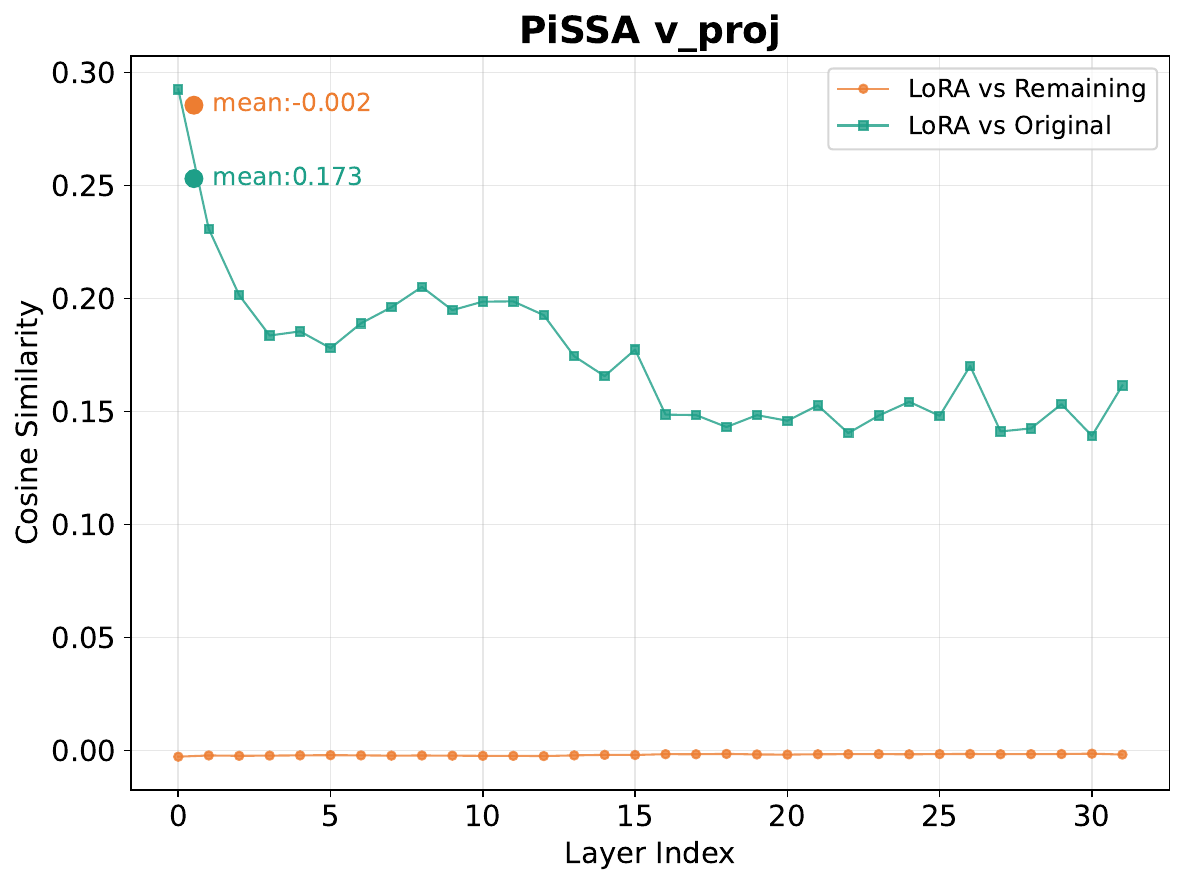}
    \end{subfigure}
    \hfill
    \begin{subfigure}[b]{0.23\textwidth}
        \centering
        \includegraphics[width=\textwidth]{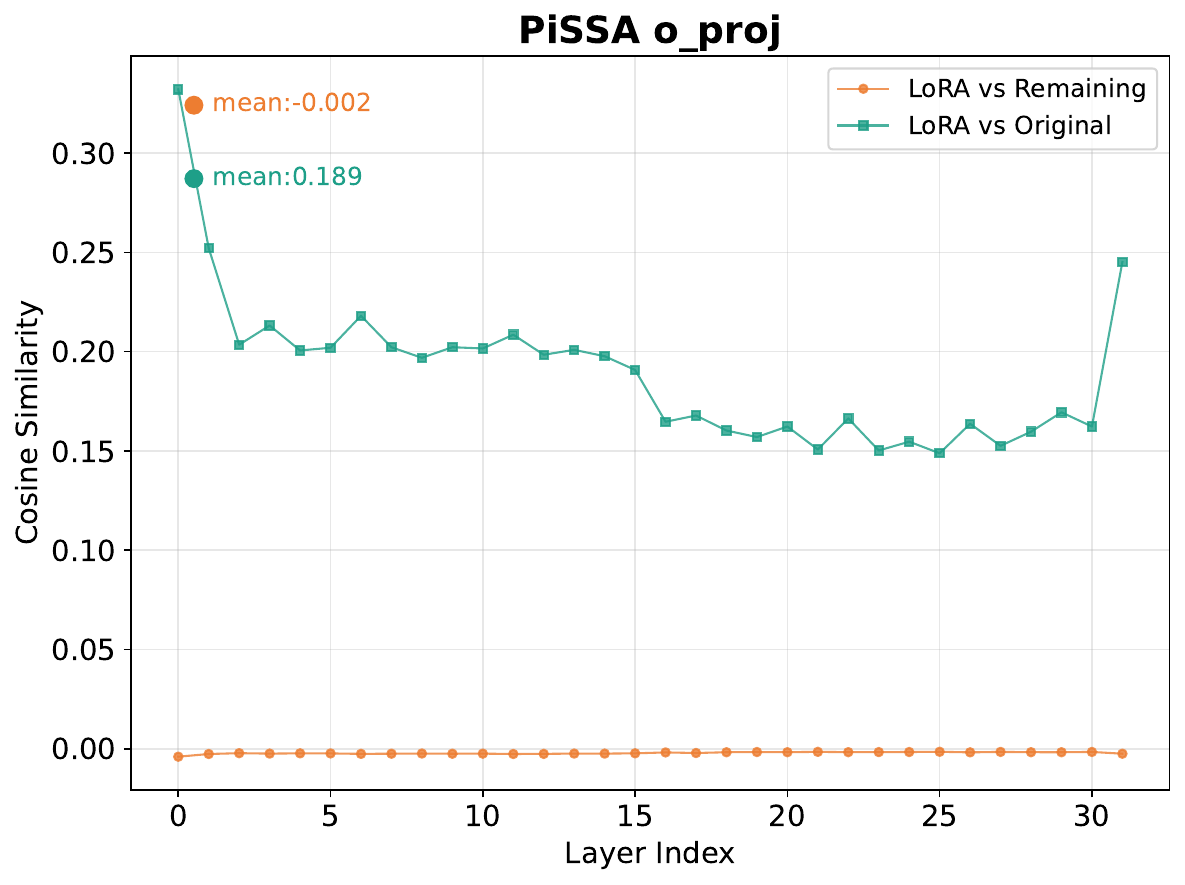}
    \end{subfigure}
    \\
    \hfill
    \begin{subfigure}[b]{0.23\textwidth}
        \centering
        \includegraphics[width=\textwidth]{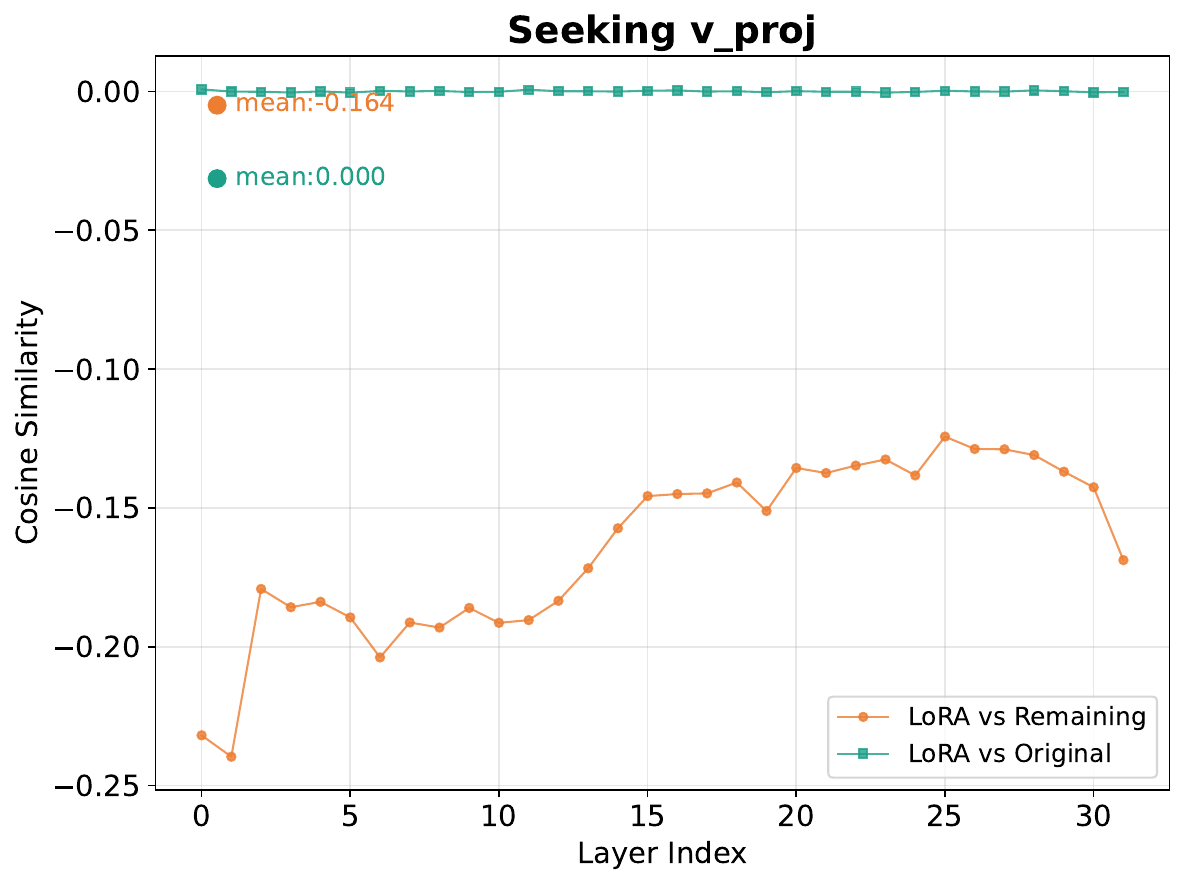}
    \end{subfigure}
    \hfill
    \begin{subfigure}[b]{0.23\textwidth}
        \centering
        \includegraphics[width=\textwidth]{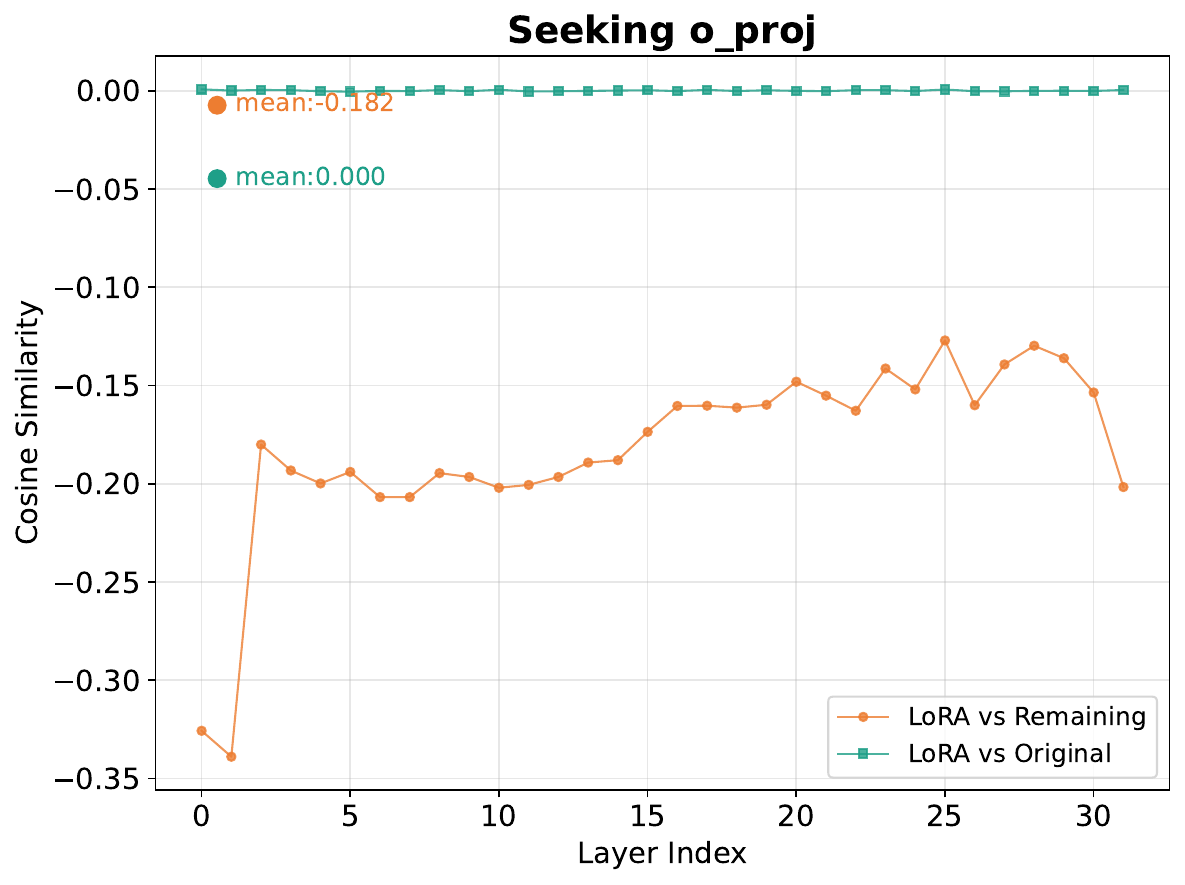}
    \end{subfigure}

        \caption{Results for Parametric Similarity Comparison between LLMs in MHSA Modules.}
    \label{fig:parametric_similarity_mhsa}
\end{figure}

\noindent \textbf{Parametric Similarity in MHSA Modules.} As shown in Figure~\ref{fig:parametric_similarity_mhsa}, we observe a similar pattern in the MHSA modules. This result indicates that the similarity between $W^l$ and $W^l_{\textrm{LoRA}}$ plays a crucial role in subsequent SFT.


\paragraph{\noindent \textbf{Comparison with Language-based Knowledge Distillation.}}

To further assess the effectiveness of LaTen in low-resource settings, we expanded our experiments to include comparisons between Llama-2-7B-Chat and Llama-2-13B-Chat, as well as additional evaluations on the Qwen2.5 models. The experimental results are summarized in Tables~\ref{tab:compare_distill_llama} and~\ref{tab:compare_distill_qwen}. These findings demonstrate that LaTen consistently outperforms distillation baselines in scenarios with extremely limited training data. Specifically, LaTen achieves performance that is comparable to or better than approaches such as supervised knowledge distillation (Supervised KD)\cite{hinton2015distilling}, sequential knowledge distillation (SeqKD)\cite{kim2016sequence}, and generalized knowledge distillation (GKD)~\cite{agarwal2024policy}, while requiring significantly fewer training examples.

These results underscore the potential of LaTen for efficient parametric knowledge transfer, particularly in low-resource scenarios where access to training data is highly constrained. However, its performance remains limited by certain vulnerabilities, suggesting that exploring more robust methods for parametric knowledge transfer presents an intriguing direction for future research.

\begin{table}[h!]
\centering
\resizebox{1.0\columnwidth}{!}{
\begin{tabular}{lcc}
\toprule
\textbf{Models} & \textbf{GSM8K} & \textbf{Traning Data} \\
\hline
Llama-2-7B-Chat & 16.07  & -\\
Llama-2-13B-Chat & 20.55 & -\\
\rowcolor{gray!20} \multicolumn{3}{l}{\textit{\# Pre-Align Parametric Knowledge Transfer}}\\
Pre-Align on $\mathcal{D}_{\text{align}}^\mathcal{T}$ \\ 
\quad -LaTen  + Pre-Aligned 13B Param. (5 Steps) & 20.47 & 5$\times$16 \\
\rowcolor{gray!20} \multicolumn{3}{l}{\textit{\#  Language-based Knowledge Distillation}}\\
Distillation on $\mathcal{D}_{\text{align}}^\mathcal{T}$ \\
\quad -SeqKD (5 Epochs) & 16.60  & 5$\times$80\\
\quad -Supervised KD (5 Epochs) & 16.60& 5$\times$80 \\
\quad -GKD (5 Epochs) & 16.60 & 5$\times$80\\
\bottomrule
\end{tabular}
}
\caption{Results for comparison with language-based knowledge distillation baselines in the Llama-2 series. Both Pre-Align (step size 16) and distillation are conducted on $\mathcal{D}_{\text{align}}^\mathcal{T}$, consisting of 80 examples.}
\label{tab:compare_distill_llama}
\end{table}

\begin{table}[h!]
\centering
\resizebox{1.0\columnwidth}{!}{
\begin{tabular}{lcc}
\toprule
\textbf{Models} & \textbf{GSM8K} & \textbf{Training Data} \\
\hline
Qwen2.5-1.5B & 62.02 & - \\
Qwen2.5-3B & 73.24  & -\\

\rowcolor{gray!20} \multicolumn{3}{l}{\textit{\# Pre-Align Parametric Knowledge Transfer}}\\
Pre-Align on $\mathcal{D}_{\text{align}}^\mathcal{T}$ \\ 
\quad -LaTen  + Pre-Aligned 3B Param. (5 Steps) & 62.32  & 5$\times$16\\
\rowcolor{gray!20} \multicolumn{3}{l}{\textit{\#  Language-based Knowledge Distillation}}\\
Distillation on $\mathcal{D}_{\text{train}}^\mathcal{T}$ \\
\quad -SeqKD (5 Epochs) & \textbf{62.47} & 5$\times$1000  \\
\quad -Supervised KD (5 Epochs) & 61.87   & 5$\times$1000\\
\quad -GKD (1 Epoch) & 62.02  & 1$\times$1000 \\
\bottomrule
\end{tabular}
}
\caption{Results for comparison with language-based knowledge distillation baselines in the Qwen2.5 series. Both Pre-Align (step size 16) and distillation are conducted on $\mathcal{D}_{\text{train}}^\mathcal{T}$, consisting of 1000 examples.}
\label{tab:compare_distill_qwen}
\end{table}

\end{document}